\documentclass[journal]{IEEEtran}

\usepackage{amsmath,amsfonts}
\usepackage{algorithmic}
\usepackage{algorithm}
\usepackage{array}
\usepackage[caption=false,font=normalsize,labelfont=sf,textfont=sf]{subfig}
\usepackage{textcomp}
\usepackage{stfloats}
\usepackage{url}
\usepackage{verbatim}
\usepackage{graphicx}
\usepackage{cite}
\usepackage{multirow}
\usepackage{color}
\usepackage{booktabs}
\usepackage{xcolor}
\usepackage{colortbl}

\usepackage{amsthm,amssymb}
\usepackage{mathrsfs}

\begin{document}

{
\twocolumn

\title{HGFormer: Topology-Aware Vision Transformer with HyperGraph Learning}

\author{Hao Wang, Shuo Zhang, Biao Leng\IEEEauthorrefmark{1}

\thanks{This work was supported by the project of the State Key Laboratory of Software Development Environment under Grant SKLSDE-2023ZX-18.}
\thanks{Hao Wang and Biao Leng are with School of Computer Science and Engineering, Beihang University, Beijing 100191, China (e-mail: lenny@buaa.edu.cn; lengbiao@buaa.edu.cn).}
\thanks{Shuo Zhang is with Beijing Key Laboratory of Traffic Data Analysis and Mining, School of Computer Science \& Technology, Beijing 100044, China (e-mail: zhangshuo@bjtu.edu.cn).}\thanks{Corresponding author: Biao Leng (lengbiao@buaa.edu.cn)}}

\markboth{SUBMISSION TO IEEE TRANSACTIONS ON xxxxxxx}%
{Shell \MakeLowercase{\textit{et al.}}: Bare Demo of IEEEtran.cls for IEEE Journals}


\maketitle

\begin{abstract}
The computer vision community has witnessed an extensive exploration of vision transformers in the past two years. 
Drawing inspiration from traditional schemes, numerous works focus on introducing vision-specific inductive biases.
However, the implicit modeling of permutation invariance and fully-connected interaction with individual tokens disrupts the regional context and spatial topology, further hindering higher-order modeling. 
This deviates from the principle of perceptual organization that emphasizes the local groups and overall topology of visual elements.
Thus, we introduce the concept of hypergraph for perceptual exploration.
Specifically, we propose a topology-aware vision transformer called HyperGraph Transformer (HGFormer).
Firstly, we present a Center Sampling K-Nearest Neighbors (CS-KNN) algorithm for semantic guidance during hypergraph construction.
Secondly, we present a topology-aware HyperGraph Attention (HGA) mechanism that integrates hypergraph topology as perceptual indications to guide the aggregation of global and unbiased information during hypergraph messaging.
Using HGFormer as visual backbone, we develop an effective and unitive representation, achieving distinct and detailed scene depictions.
Empirical experiments show that the proposed HGFormer achieves competitive performance compared to the recent SoTA counterparts on various visual benchmarks. 
Extensive ablation and visualization studies provide comprehensive explanations of our ideas and contributions.
\end{abstract}


\section{Introduction}
\label{intro}
\IEEEPARstart{T}{he} introduction of transformers to vision has shown remarkable performance on various visual tasks~\cite{vit,carion2020end}. Built upon self attention, vision transformers allow all tokens to communicate with a weak inductive bias, capturing long-range dependencies~\cite{raghu2021vision}, achieving universal~\cite{lu2022frozen}, flexible~\cite{naseer2021intriguing}, robust~\cite{shao2021adversarial} modeling capabilitiy and scalability~\cite{zhai2022scaling}. 
However, the paradigm of implicit modeling is uninformative and inefficient. Previous studies show that the architecture lacks an intrinsic inductive bias in local structure and spatial correlation~\cite{zhang2023vitaev2,lu2022bridging,wei2023sparsifiner}. To verify the hypotheses, we conduct a study through visualization and find that this leads to ambiguous and chaotic scene understanding in Fig.~\ref{fig:feat_intro}b.

The permutation invariance and the fully-connected interaction with individual tokens in vision transformer disrupt the regional context and spatial topology, further hindering higher-order modeling. 
Instead of the paradigm, one appealing idea is to organize and exploit the holistic structure of environment. 
This thought starts from the principle of perceptual organization~\cite{lowe2012perceptual} that emphasizes structuring individual visual elements into coherent groups and establishing interactions among groups, thereby simulating human cognition in complex scenes. 
Thus, we seek to introduce perceptual indications to mitigate the ambiguity and chaos arising from the implicit modeling in vision transformer. 

\begin{figure}[t]
    \centering
    \includegraphics[width=1\columnwidth]{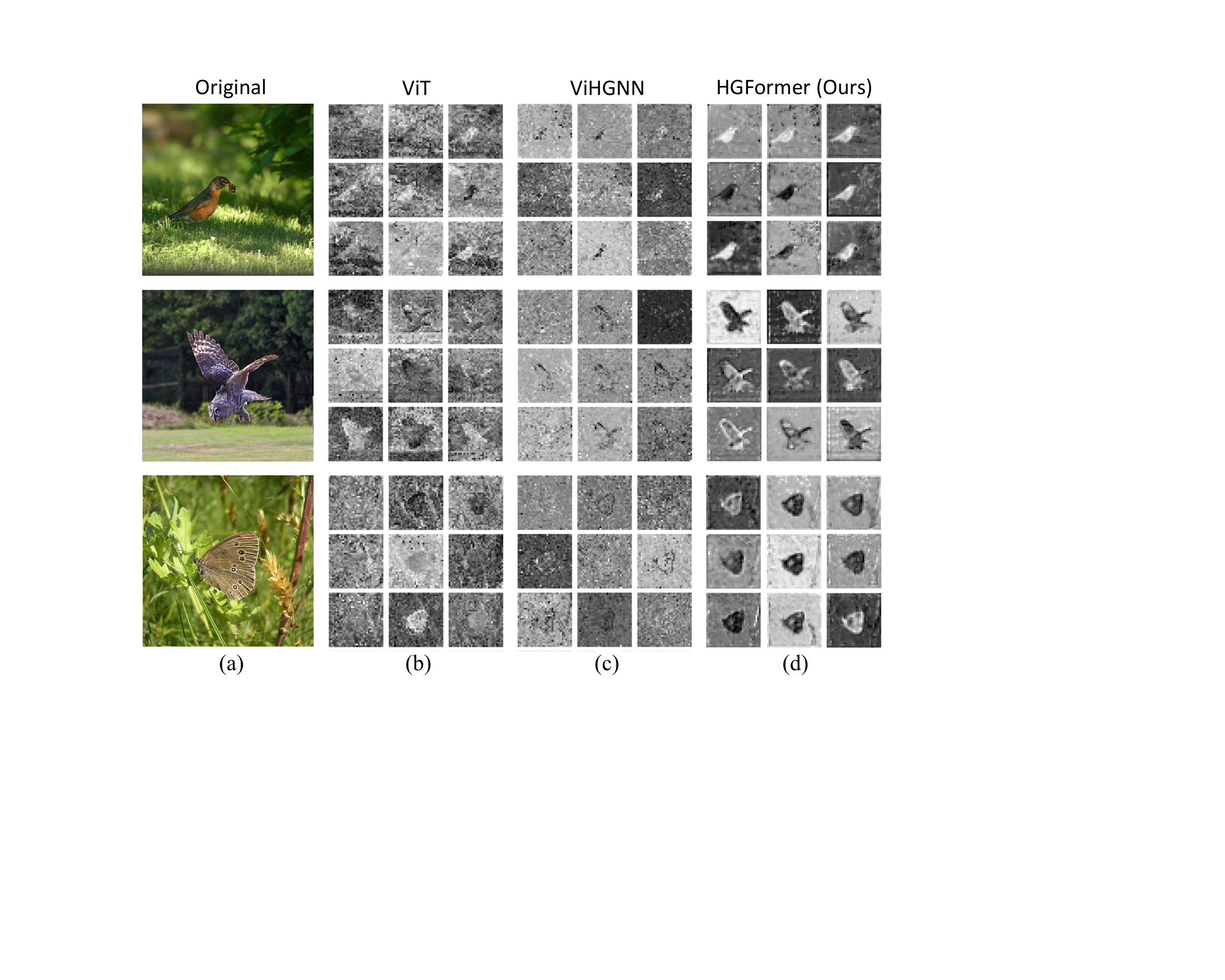}
    \vspace{-20pt}
    \caption{
    \textbf{Visualizations of feature maps in different methods on ImageNet}. 
    (a) Original Image. 
    (b) ViT. 
    (c) ViHGNN.
    (d) HGFormer(Ours).
    ViT blends the foreground with the background ambiguously. 
    ViHGNN distinguishes between foreground and background, but the portrayals of objects is rough and unclear. 
    HGFormer(Ours) significantly highlights the foreground and suppresses the background, achieving a detailed depiction of objects.
    \textit{
    Zoom in for better view.
    }
    }
    \vspace{-17pt}
    \label{fig:feat_intro}
\end{figure}

The recent development of hypergraph~\cite{feng2019hypergraph,bai2021hypergraph,gao2022hgnn+} has shown promising progress, providing us an avenue for perceptual exploration. 
Hypergraph generalizes simple graph by permitting the connection of an arbitrary number of nodes through hyperedge, enabling the simulation of regional context and spatial topology. 
Hypergraph messaging with node-hyperedge-node subroutines shows strong capabilities in topology learning and higher-order modeling.
Recent ViHGNN~\cite{han2023vision} attempts to construct hypergraph based on traditional clustering algorithm and perform representation learning by stacking multiple hypergraph convolutions (HGConv)~\cite{gao2022hgnn+}.
But the randomness of its clustering algorithm is susceptible to noise or outliers.
The strict structural inductive bias of cascaded HGConvs causes locality, over-smoothing and error accumulation~\cite{sheaf,chen2022preventing,liu2021strongly}, manifesting as visually rough and unclear portrayals in Fig.~\ref{fig:feat_intro}c.

Motivated by the above analyses and observations, we propose to rethink the topology perception of HGConv and the global understanding of Transformer.
HGConv inherently integrates regional context and spatial topology within the hypergraph domain. 
In contrast, transformer impartially gathers global context and unbiased particulars from the whole domain.
Thus, we propose a topology-aware HyperGraph Attention (HGA) that incorporates hypergraph topology as perceptual indications to guide the aggregation of global and unbiased information. 
Firstly, we employ HGConv to learn the hypergraph local topologies and predict the hypergraph tokens as perceptual indications.
Then, we employ the generated hypergraph tokens to guide the aggregation of global context and unbiased particulars for contextual refinement.
As visualized in Fig.~\ref{fig:feat_intro}d, we develop an effective and unitive representation, achieving distinct and detailed scene depictions.

\begin{figure*}[!t]
    \centering
    \includegraphics[width=0.98\textwidth]{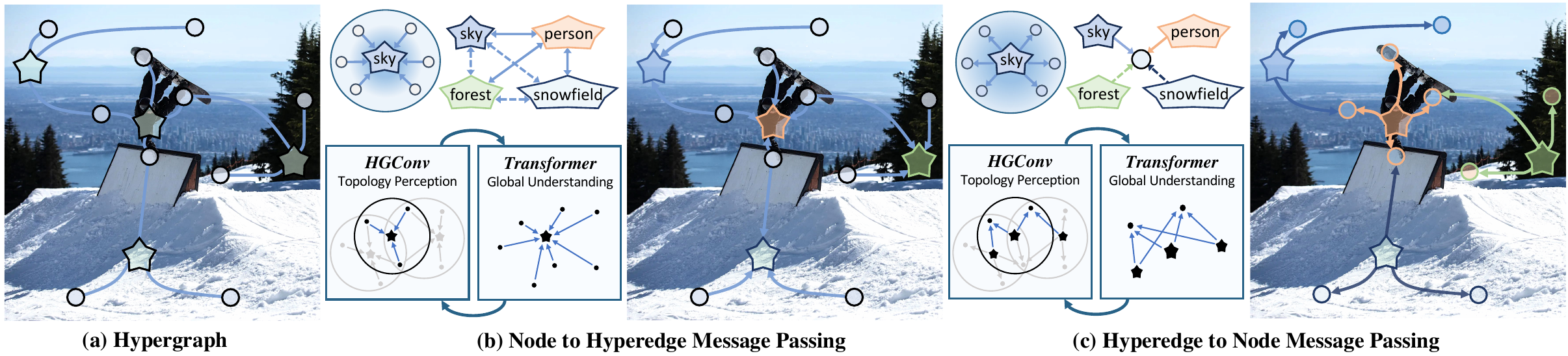}
    \vspace{-8pt}
    \caption{
      \textbf{Illustration of hypergraph concept.} 
      (a) Moving beyond grid or sequence, input feature map is transformed into hypergraph by the proposed CS-KNN algorithm, where the stars denote sampling centers and the nodes connected by them signify hyperedges with semantic dependencies.
      (b) Hyperedge tokens are generated by aggregating their relevant nodes within the hyperedge local topologies, during which higher-order semantics are explored and irrelevant noise is potentially eliminated.
      (c) Node tokens are updated by aggregating their relevant hyperedges within the node local topologies, during which message across all nodes is enhanced and propagated.
      \textit{As messaging functions, the proposed topology-aware HGA incorporates the topology perception of HGConv as perceptual indications and the global understanding of Transformer for contextual refinement.}
    \textit{
    Zoom in for better view.
    }
    }
    \vspace{-12pt}
    \label{fig:intro}
\end{figure*}
Based on the topology-aware HGA as messaging functions, we present a general visual backbone named HyperGraph Transformer (HGFormer). 
Within HGFormer, we deploy a stack of HGFormer blocks to facilitate visual representation learning. 
As shown in Fig.~\ref{fig:intro}, the block is configured with hypergraph messaging for relational reasoning, involving node-hyperedge-node subroutines.
Besides, to mitigate the sensitivity within traditional clustering algorithms when dealing with noise and outliers during hypergraph construction, we propose a Center Sampling K-Nearest Neighbors (CS-KNN) algorithm, which integrates the class token~\cite{liang2022not,fayyaz2022adaptive} in vision transformer for semantic guidance in clustering. 
These clusters are treated as hyperedge topologies and composed into the hypergraph topology, serving as the input for HGFormer blocks.

In summary, our contributions are as follows:

\begin{itemize}
\item We present a topology-aware HyperGraph Attention (HGA) mechanism that integrates hypergraph topology as perceptual indications to guide the aggregation of global and unbiased information during hypergraph messaging.
\item We present a Center Sampling K-Nearest Neighbors (CS-KNN) algorithm for semantic guidance during hypergraph construction.
\item We present a general visual backbone named HyperGraph
Transformer (HGFormer). Empirical experiments show that the proposed HGFormer consistently outperforms the recent SoTA counterparts across various benchmarks. Extensive ablation and visualization studies provide comprehensive explanations of our ideas and contributions.
\end{itemize}

\section{Related Work}
\label{related}
In this section, we briefly review some vision transformers, recent advancements in hypergraph learning and graph-based neural networks in vision.

\subsection{Vision Transformer}
In the past two years, the community has witnessed an extensive exploration of vision transformers. Drawing inspiration from traditional representation schemes, numerous works focus on introducing vision-specific inductive biases, such as locality, translation invariance, and 2D nature, as in convolution. 
Previous works include local attention~\cite{li2021localvit,chen2022regionvit,hassani2023neighborhood,pan2023slide,acod}, convolution enhancement~\cite{wu2021cvt,guo2022cmt,pan2022integration}, multi-scale resolutions~\cite{wang2021pyramid,jiao2023dilateformer} and hierarchical architecture~\cite{liu2021swin, han2021transformer,chu2021twins,wang2021pyramid}.
Targeting the properties of image tokens and attention mechanism, several works proposes sparse~\cite{wei2023sparsifiner,fayyaz2022adaptive,yin2022vit,liang2022not,meng2022adavit,amgtp} or merging~\cite{bolya2022token, zeng2022not} strategies.
In addition to the above traditional schemes, there are many other directions for improvement, such as training strategy~\cite{deit,wu2022tiny}, positional embeddings~\cite{vit,wu2021rethinking,chu2021conditional} and normalization operation~\cite{deit}.

Different from above works that primarily introduce vision-specific inductive biases, we start from perceptual organization that emphasizes the local groups and overall topology of scenes.
Following the principle, the implicit modeling of permutation invariance and fully-connected interaction in vision transformer disrupts the regional context and spatial topology, further hindering higher-order modeling~\cite{zhang2023vitaev2,lu2022bridging,wei2023sparsifiner}. Thus, we introduce the concept of hypergraph into transformer as perceptual indications for regional context, spatial topology and higher-order modeling.

\begin{figure*}[!t]
    \centering
    \includegraphics[width=0.98\textwidth]{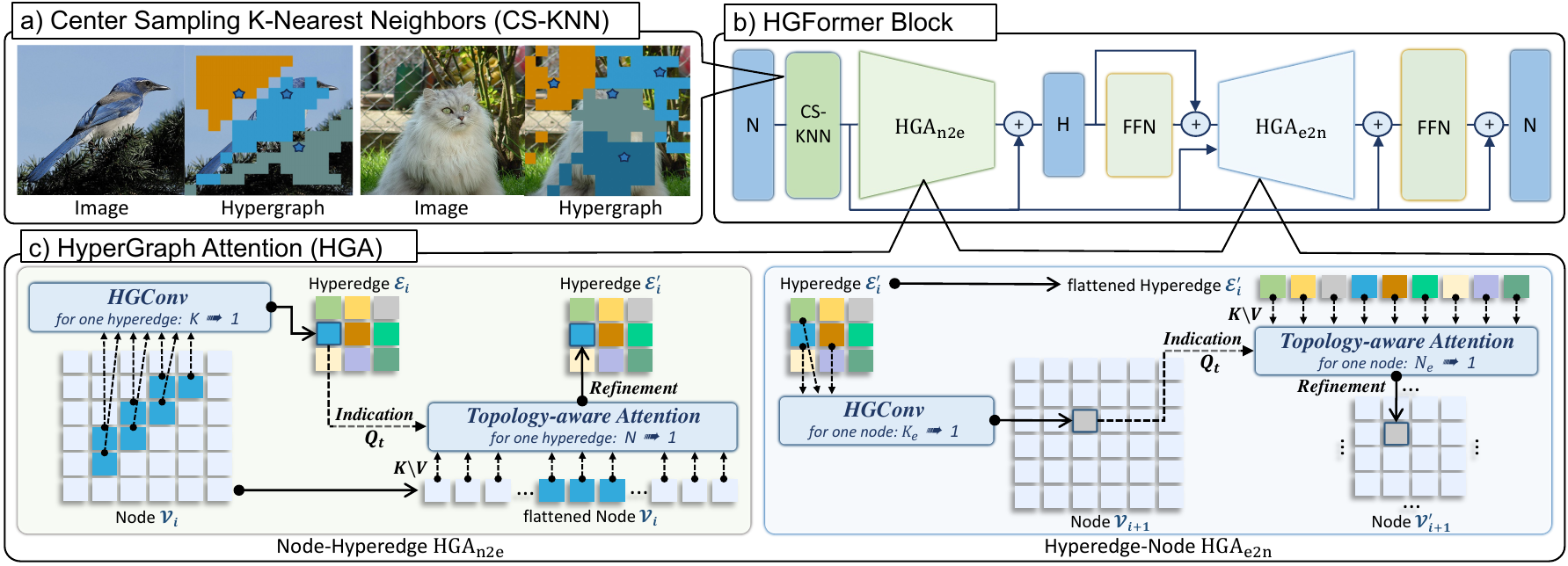}
    \vspace{-8pt}
    \caption{
    \textbf{Framework of HGFormer block.} 
    (a) Based on the proposed CS-KNN algorithm, input images are transformed into hypergraph, where the stars denote sampling centers and the nodes connected by them signify hyperedges with semantic dependencies, visualized by the same color.
    (b) HGFormer block is built based on the node-hyperedge-node messaging mechanism, enabling representation learning through the high-order relational reasoning.
    (c) As messaging functions, the proposed topology-aware HGA incorporates the topology perception of HGConv as perceptual indications and the global understanding of Transformer for contextual refinement. $K$ indicates the number of nodes within hyperedge local topology, $K_e$ indicates the number of hyperedges within node local topology, $N$ indicates the number of all nodes, $N_e$ indicates the number of all hyperedges.
    \textit{
    Zoom in for better view.
    }
    }
    \vspace{-10pt}
    \label{fig:pipline}
\end{figure*}

\subsection{Hypergraph Learning}
Hypergraph is a generalization of simple graph, and hypergraph learning has gained increasing attention for its effectiveness in higher-order modeling. Numerous hypergraph learning frameworks have been proposed~\cite{feng2019hypergraph, wang2023equivariant}. 
Following spectral graph convolution~\cite{semi, graphfilter, gdl}, Hypergraph Neural Network (HGNN)~\cite{feng2019hypergraph} first generalizes convolution operation to hypergraph learning, which is realized on spectral domain with hypergraph laplacian and truncated chebyshev polynomials.
Drawing inspiration from spatial graph convolution~\cite{diffconv,gat}, HGNN+~\cite{gao2022hgnn+} develop a general two-stage hypergraph convolution in the spatial domain.
Advanced beyond ViG, Vision HGNN (ViHGNN)~\cite{han2023vision}
attempts to model dependencies by stacking multiple hypergraph convolutions. 
Moreover, several excellent operators and stronger proofs about hypergraph~\cite{coupette2023ollivier, wang2023equivariant} have emerged and been applied in various tasks, such as community detection \cite{zhang2022sparse}, item recommendation~\cite{ding2023hyperformer} and image retrieval \cite{zeng2023multi}, showing significant performances.

Our work belongs to the spatial-based (hyper)graph convolution.
In computer vision, global understanding has proven to be advantageous~\cite{ding2022scaling, slak}. 
Existing HGNNs~\cite{feng2019hypergraph}, attention-based HGNNs~\cite{bai2021hypergraph} and ViHGNN~\cite{han2023vision} only conduct messaging according to the hypergraph local topologies.
Thus, we propose to employ transformer to relax the constraints of fixed topology, overcoming the bottleneck of strict inductive bias and achieving global modeling.
Besides, to mitigate the sensitivity within traditional clustering algorithms when dealing with noise and outliers during hypergraph construction, we propose to employ the class token~\cite{liang2022not,fayyaz2022adaptive,yin2022vit} in vision transformer for semantic guidance in clustering.

\subsection{Graph in Computer Vision.}
Convolution neural networks (ConvNets) is limited by a finite receptive field, making it inexact to capture long-range dependencies.
Previous works include convolution variants~\cite{yu2016multi,yu2017dilated,dai2017deformable,zhu2019deformable,vfigdc}, multi-scale resolutions~\cite{lin2017fpn,liu2018pan,ma2021spatial,multi-focus,extendedFPN}, attention mechanism~\cite{hou2021coordinate,pan2022integration,yang2022moat,zhang2021self,lin2021structured}, improved architecture~\cite{chollet2017xceptipn,andrew2017mobilenet} and large kernel~\cite{liu2022convnet, ding2022scaling,wang2023can}. 
Specially, there has been a specific kind of work that introduces graph representation and graph neural network (GNN) ~\cite{chen2019graph,li2020spatial,zhang2020dynamic,ding2021interaction,li2018beyond,zhai2021mutual} to optimize feature extraction format and reason relations between distant regions. 
Recently, a novel vision graph neural networks (ViG)~\cite{han2022vision} has been proposed, which treats image patches as nodes, constructs semantic graph and employs graph convolutions to extract information.
The novel architecture and its variations~\cite{munir2023mobilevig, wu2023pvg} achieve competitive results on various visual benchmarks. 
Moreover, GNNs are also popular in downstream visual tasks, 
such as action recognition~\cite{zhou2021graph}, point cloud analysis~\cite{wang2019dgcnn} and few-shot learning~\cite{ling2020dpgn}.

From the perspective of graph, transformer can also be regarded as the fully-connected graph~\cite{dwivedi2020generalization,rampavsek2022recipe}. In visual analysis, the visual objects and scenes often consist of multiple tokens, which are difficultly simulated by the pairwise connections in graph or transformer. Thus, we introduce the concept of hypergraph to compensate for the multiple dependencies and higher-order modeling.

\section{Method}
\subsection{Overview}
The details of HyperGraph Transformer (HGFormer) block are shown in Fig.~\ref{fig:pipline}, consisting of two steps: hypergraph construction and hypergraph messaging. 
The former transforms the input image into hypergraph topology by the proposed Center Sampling K-Nearest Neighbors (CS-KNN) algorithm. 
The latter adopts the proposed node-hyperedge-node HyperGraph Attention (HGA) for representation learning.

\subsubsection{Hypergraph Construction} We first transform the input image into its hypergraph topology. Generally, the hypergraph is defined as $\mathcal{HG} = (\mathbb{V}, \mathbb{E})$, where $\mathbb{V}$ denotes the node set and $\mathbb{E}$ denotes the hyperedge set. 
It generalizes simple graph by permitting the connection of an arbitrary number of nodes through hyperedge, enabling the simulation of regional context and spatial topology. 
The objective of hypergraph construction is to create a hypergraph incidence matrix $\mathbb{H}$ that encodes the connection between the node set $\mathbb{V}$ and the hyperedge set $\mathbb{E}$, as a mathematical representation of the hypergraph topology.

For an input image, pixel tokens $\mathcal{X}_i \in \mathbb{R}^{N \times C}$ at the $i^{th}$ layer ($N = W \times H$ is pixel number, $C$ is embedding dimension) are treated as node tokens $\mathcal{V}_i \in \mathbb{R}^{N \times C}$. Subsequently, the hypergraph incidence matrix $\mathcal{H}_i$ is created through the proposed CS-KNN algorithm:
\begin{equation}
\begin{split}
\label{eq:over1}
\mathcal{H}_i & = \mathrm{CS{-}KNN}(\mathcal{V}_i) \\
\end{split}
\end{equation}
where $\mathcal{H}_i \in \{0,1\}^{N \times N_e}$ ($N$ is node number, $N_e$ is hyperedge number) denotes the constructed hypergraph incidence matrix. 
This matrix, as the representation of the hypergraph topology $\mathcal{HG}$, directs the subsequent messaging computations.
Specifically, CS-KNN algorithm incorporates semantic guidance during hypergraph construction for robust hypergraph topologies.

\subsubsection{Hypergraph Messaging} For respresentation learning on the constructed hypergraph $\mathcal{HG}$, the HGFormer block is configured with hypergraph messaging for relational reasoning, involving node-$\mathbb{V}$ to hyperedge-$\mathbb{E}$ to node-$\mathbb{V}$ subroutines. 
In the first stage, hyperedge tokens are generated by aggregating their relevant nodes within the hyperedge local topologies, whereby higher-order semantics are explored and irrelevant noise is potentially eliminated.
In the second stage, node tokens are updated by aggregating their relevant hyperedges within the node local topologies, during which message across all nodes is enhanced and propagated.
Therefore, the bi-directional hypergraph messaging aims to enhance the capabilities of topology learning and higher-order modeling. 

Specifically, given the node tokens $\mathcal{V}_i$ and the hypergraph topology $\mathcal{H}_i$, the whole messaging is formulated by a node-hyperedge HyperGraph Attention ($\mathrm{HGA_{n2e}}$) and a hyperedge-node HyperGraph Attention ($\mathrm{HGA_{e2n}}$) as follows:
\begin{equation}
\begin{split}
\label{eq:over2}
\mathcal{E}_i' & = \mathrm{HGA_{n2e}}(\mathcal{V}_i,\mathcal{H}_i) \\
\mathcal{V}_{i+1}' & = \mathrm{HGA_{e2n}}(\mathcal{E}_i',\mathcal{H}_i) \\
\end{split}
\end{equation}
where $\mathcal{E}_i' \in \mathbb{R}^{N_e \times C}$ denotes the generated hyperedge tokens and $\mathcal{V}_{i+1}' \in \mathbb{R}^{N \times C}$ denotes the updated node tokens for the ${i+1}^{th}$ layer. 
Specifically, in $\mathrm{HGA_{n2e}}$, message runs along the hypergraph topology $\mathcal{H}_i$ from node $\mathcal{V}_i$ to hyperedge $\mathcal{E}_i'$. In $\mathrm{HGA_{e2n}}$, message runs along the hypergraph topology $\mathcal{H}_i$ from hyperedge $\mathcal{E}_i'$ to node $\mathcal{V}_{i+1}'$.
In summary, the whole messaging is oriented by hypergraph topology $\mathcal{H}_i$ and mediated through hyperedge tokens $\mathcal{E}_i'$, reasoning about correlations and distilling characteristics from a higher-order perspective. 

\subsection{Center Sampling K-Nearest Neighbors (CS-KNN)}
\label{Hypergraph Construction}
For hypergraph construction, clustering algorithms~\cite{gao2022hgnn+, han2023vision} have been the most widespread. But its randomness makes it vulnerable to noise or outliers, which is especially problematic in situations where images often contain excessive noise and isolated pixels.
To obtain robust hypergraph topologies, we propose a Center Sampling K-Nearest Neighbors (CS-KNN) algorithm for semantic guidance. 
Specifically, inspired by~\cite{liang2022not,fayyaz2022adaptive}, we utilize the class token to identify the informative tokens as sampling cluster centers.

\subsubsection{Center Sampling} 
For an input image, pixel tokens $\mathcal{X}_i \in \mathbb{R}^{N \times C}$ are treated as node tokens $\mathcal{V}_i \in \mathbb{R}^{N \times C}$.
To sample informative nodes with semantics, we compute attentiveness of the class token $x^{cls}_i \in \mathbb{R}^{1 \times C}$ with respect to each node token $x_i \in \mathbb{R}^{1 \times C}$ to determine its importance score, which is formulated by normalized dot product of the two token vectors as follows:
\begin{equation}
    \begin{split}
    \label{eq:dist1}
        \mathrm{Score}(x^{cls}_i, x) = (x^{cls}_i \cdot x_i) /\sqrt{d}
    \end{split}
\end{equation}
Here, $x^{cls}_i$ is the class token, $x_i$ is node token and $d$ is vector dimension. 
For node tokens with high scores, it indicates that the tokens are semantically representative and informative.
We select $N_e$ tokens with the highest attention values as cluster centers $\mathcal{X}^{ctr}_i \in \mathbb{R}^{N_e \times C}$ ($N_e$ is hyperedge number).

\subsubsection{K-Nearest Neighbor} 
For clustering based on the sampling centers, we measure semantic distances between center $x^{ctr}_i \in \mathbb{R}^{1 \times C}$ and node token $x_i \in \mathbb{R}^{1 \times C}$ by normalized dot product of the two token vectors as follows:
\begin{equation} 
    \begin{split}
    \label{eq:dist2}
        \mathrm{Distance}(x^{ctr}_i, x) = (x^{ctr}_i \cdot x_i) /\sqrt{d}
    \end{split}
\end{equation}
Here, $x^{ctr}_i$ is the sampling cluster center, $x_i$ is node token and $d$ is vector dimension. 
Each cluster center $x^{ctr}_i$ takes the k nearest tokens as its neighbors and merges as a cluster. Given all the clusters, we treat them as hyperedge topologies and compose them into the final hypergraph topology, mathematically represented as the hypergraph incidence matrix $\mathcal{H}_i \in \{0,1\}^{N \times N_e}$. 
The computations of Eq.~\ref{eq:dist1} and Eq.~\ref{eq:dist2} conform to the normalized dot product in~\cite{clip}.

Although grouping or clustering strategies~\cite{zeng2022not, groupvit, clip} have been widely employed in current vision or multimodal transformers, an issue arises by the existence of noise, outliers, or uninformative tokens within text or images.
Existing methods indiscriminately employ each token as a center for group formation, consequently leading to random and confused groups, exemplified in Fig.~\ref{fig:knn}.
In contrast and as an advancement, the proposed CS-KNN utilizes the class token to identify informative tokens as sampling centers, filtering out a substantial amount of noise and outliers to construct robust hyperedge groups.
The quantitative comparisons in Tab. \ref{tab:algo} show that the proposed CS-KNN algorithm performs better.

\subsection{Node-Hyperedge HyperGraph Attention ($HGA_{n2e}$)}
At this stage, message runs along the hypergraph topology $\mathcal{H}_i$ from node $\mathcal{V}_i$ to hyperedge $\mathcal{E}_i'$, whereby higher-order semantics are explored and irrelevant noise is potentially eliminated. 
Specifically, we present a topology-aware $\mathrm{HGA_{n2e}}$ to learn about the generation and refinement of hyperedge tokens. 
Following spatial-based graph convolution~\cite{diffconv,gat,gao2022hgnn+}, 
given the hypergraph incidence matrix $\mathcal{H}_i$ and node tokens $\mathcal{V}_i$, hypergraph convolution (HGConv) first learns sparse weights about node tokens $\mathcal{V}_i$ within the hyperedge topologies of $\mathcal{H}_i$ to predict hyperedge tokens $\mathcal{E}_i$:
\begin{equation} 
    \begin{split}
    \label{eq:v2e_conv}
        \mathcal{E}_i = \sigma (\mathcal{W} \mathcal{D}_{e}^{-1} \mathcal{H}^T_i \mathcal{V}_i) \\
    \end{split}
\end{equation}

Here, $\mathcal{E}_i \in \mathbb{R}^{N_e \times C}$ is the generated hyperedge tokens with information of hyperedge topologies and corresponding regional context. 
$\sigma(\cdot)$ is activation function, $\mathcal{W} \in \mathbb{R}^{C \times C}$ is learnable mapping weights, $\mathcal{D}_e \in \mathbb{R}^{N_e \times N_e}$ is degree matrix of hyperedge topologies and $\mathcal{H}^T_i \in \mathbb{R}^{N_e \times N}$ is transpose of hypergraph incidence matrix.
At this point, hyperedge tokens $\mathcal{E}_i$ integrates spatial and regional context within the hyperedge topologies.

Subsequently, we employ the hyperedge tokens $\mathcal{E}_i$ as perceptual indications to guide the aggregation of global context and unbiased particulars from the whole node tokens $\mathcal{V}_i$ through the proposed topology-aware attention.
Specifically, the hyperedge tokens $\mathcal{E}_i$ are served as topology-aware query $Q_t$, and the node tokens $\mathcal{V}_i$ are served as key $K$ and value $V$:
\begin{equation} 
    \begin{split}
    \label{eq:v2e-attn}
        Q_t & = \mathcal{E}_iW^q, K = \mathcal{V}_iW^k, V = \mathcal{V}_iW^v \\
        \mathcal{E}_i' & = \mathrm{Softmax}(Q_tK^T/\sqrt{d_k})V \\
    \end{split}
\end{equation}
Here, $\mathcal{E}_i' \in \mathbb{R}^{N_e \times C}$ denotes renewed hyperedge tokens with contextual refinement. $W^q,W^k,W^v$ are all projection weight and $d_k$ is vector dimension. We adopt the multi-head settings and feedforward modules to explore the abstract and diverse representation of hyperedge tokens. 
In summary, the renewed hyperedge tokens $\mathcal{E}_i'$ have achieved a unitive representation, taking into account both the hyperedge topologies and global understanding.

\subsection{Hyperedge-Node HyperGraph Attention ($HGA_{e2n}$)}

\begin{figure*}[!t]
    \centering
    \includegraphics[width=0.98\textwidth]{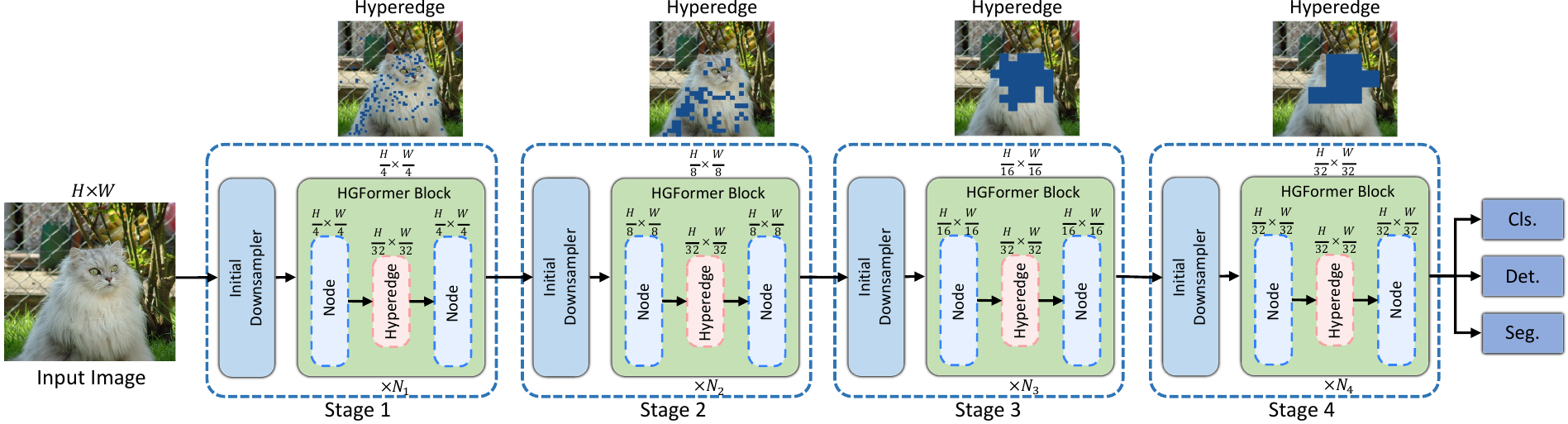}
    \vspace{-9pt}
    \caption{
    \textbf{Architecture of HGFormer network with four stages.} Following~\cite{he2016deep,wang2021pyramid}, HGFormer network is constructed as a 4-stage pyramid architecture. Each stage $i$ consists of an embedding module and $N_i$ HGFormer blocks.
    Within HGFormer block, the input token map is transformed into hypergraph, where nodes with dependencies are assigned into the same hyperedge, visualized on the upper part.
    HGFormer block is configured with node-hyperedge-node messaging mechanism for relational reasoning.
    \textit{
    Zoom in for better view.
    }
    }
    \vspace{-10pt}
    \label{fig:architecture}
\end{figure*}

At this stage, message runs along the hypergraph topology $\mathcal{H}_i$ from hyperedge $\mathcal{E}_i'$ to node $\mathcal{V}_{i+1}'$ and message among nodes is propagated through related hyperedges. 
Specifically, we present a topology-aware $\mathrm{HGA_{e2n}}$ to learn about the generation and refinement of node tokens.
Given the hypergraph incidence matrix $\mathcal{H}_i$ and hyperedge tokens $\mathcal{E}_i'$, HGConv first learns sparse weights about hyperedge tokens $\mathcal{E}_i'$ within the node topologies of $\mathcal{H}_i$ to predict node tokens $\mathcal{V}_{i+1}$:
\begin{equation} 
    \begin{split}
    \label{eq:e2v_conv}
        \mathcal{V}_{i+1} = \sigma (\mathcal{W} \mathcal{D}_{v}^{-1} \mathcal{H}_{i} \mathcal{E}_i') \\
    \end{split}
\end{equation}
Here, $\mathcal{V}_{i+1} \in \mathbb{R}^{N \times C}$ is the generated node tokens with information of node topologies and corresponding regional context. $\sigma(\cdot)$ is an activation function, $\mathcal{W} \in \mathbb{R}^{C \times C}$ is learnable mapping weight, $\mathcal{D}_v \in \mathbb{R}^{N \times N}$ is degree matrix of node topologies and $\mathcal{H}_{i} \in \mathbb{R}^{N \times N_e}$ is hypergraph incidence matrix. 
At this point, node tokens $\mathcal{V}_{i+1}$ integrates spatial and regional context within the node topologies.

Subsequently, we employ the node tokens $\mathcal{V}_{i+1}$ as perceptual indications to guide the aggregation of global context and unbiased particulars from the whole hyperedge tokens $\mathcal{E}_i'$ through the proposed topology-aware attention. 
Specifically, the node tokens $\mathcal{V}_{i+1}$ are served as topology-aware query $Q_t$, and the hyperedge tokens $\mathcal{E}_i'$ are served as key $K$ and value $V$:
\begin{equation} 
    \begin{split}
    \label{eq:e2v-attn}
        & Q_{t}  = \mathcal{V}_{i+1}W^q, K = \mathcal{E}_i'W^k, V = \mathcal{E}_i'W^v \\
        & \mathcal{V}_{i+1}' = \mathrm{Softmax}(Q_tK^T/\sqrt{d_k})V \\
    \end{split}
\end{equation}
Here, $\mathcal{V}_{i+1}' \in \mathbb{R}^{N \times C}$ denotes refined node tokens with contextual refinement. $W^q,W^k,W^v$ are all projection weight and $d_k$ is vector dimension. We adopt the multi-head settings and feedforward modules to enhance diversity of node tokens $\mathcal{V}_{i+1}'$ and address the over-smoothing problem.
In summary, the refined node tokens $\mathcal{V}_{i+1}'$ have achieved a unitive representation, taking into account both the node topologies and global understanding.
Additionally, we replace fixed-size positional encoding with convolutional feedforward modules. 

\section{HGFormer Network} 
Based on the above HGFormer block, we construct a 4-stage pyramid HGFormer network as visual backbone, illustrated in Fig.~\ref{fig:architecture}. 
Within each stage, the feature map is first fed into an embedding module to adjust its resolution and dimension. Then, the embedded tokens flow into several HGFormer blocks.
Specifically, resolutions are reduced into $\frac{H}{4}\times\frac{W}{4}$ $\frac{H}{8}\times\frac{W}{8}$, $\frac{H}{16}\times\frac{W}{16}$, $\frac{H}{32}\times\frac{W}{32}$ and dimensions are expended into $C$, $2C$, $5C$ and $8C$ at each stage, respectively.

\setlength{\tabcolsep}{11pt}
\begin{table}[t]
    \centering
    \caption{\textbf{HGFormer variants.} 
    Detailed configurations of HGFormer networks. \textbf{`T'} denotes tiny, \textbf{`S'} denotes small and \textbf{`B'} denotes base. For all networks, the dimension per head is 32 and the expansion ratio of MLP is 4. 
    }
    \vspace{-5pt}
    \resizebox{0.47\textwidth}{!}{
    \begin{tabular}{l|c|c|c|c}
        \toprule
        \textbf{Variant} & \textbf{Channels}  & \textbf{Stages} & \textbf{Params} & \textbf{FLOPs} \\
        \midrule
        \textbf{HGFormer-T}     & 32        & 1, 2, 4, 2        &  5 M          & 1.2 G \\
        \textbf{HGFormer-S}     & 64        & 1, 2, 4, 2        &  28 M         & 4.5 G \\
        \textbf{HGFormer-B}     & 96        & 1, 2, 4, 2        &  54 M         & 8.7 G \\
        \bottomrule
    \end{tabular}
    }
    \vspace{-12pt}
    \label{tab:variants}
\end{table}

Within HGFormer block, input feature map is first transformed into hypergraph topology.
Following the spatial reduction mechanism in~\cite{wang2021pyramid, wang2022pyramid2}, the reduction ratio of the hyperedge number $N_e$ at each stage is set as [$\frac{1}{8}$,$\frac{1}{4}$,$\frac{1}{2}$,$1$] by the proposed center sampling strategy to reduce computation and redundancy. 
We also illustrate the evolution of hypergraph at four stages, where the masks represent hyperedge topology and the corresponding regions share dependencies. 
In the shallow stage, node selection is based on low-level characters such as color and texture. 
As the stages deepen, node selection becomes increasingly inclined towards semantics.
The proposed hypergraph topology offers information of local context and spatial topology.
For comprehensive evaluation and comparison, we build three HGFormer variants with different configurations in Tab.~\ref{tab:variants}. 

\section{Experiments}

\begin{figure}[t]
    \centering
    \includegraphics[width=1\columnwidth]{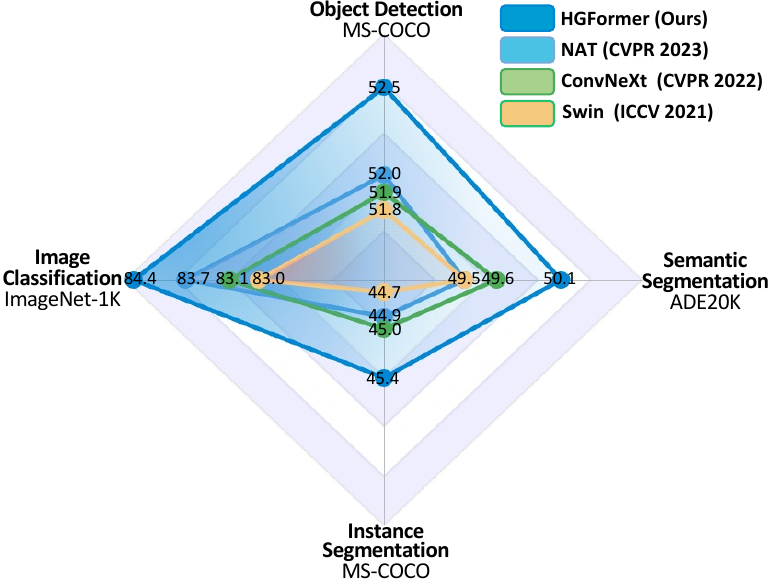}
    \vspace{-20pt}
    \caption{
    \textbf{Performance Comparisons between the proposed method and the recent SoTA models~\cite{hassani2023neighborhood,liu2022convnet,liu2021swin}.} Our work outperforms these Transformer and ConvNet counterparts with similar parameters and computation.
    }
    \vspace{-14pt}
    \label{fig:performance}
\end{figure}

We compare the proposed HGFormer with the recent SoTA counterparts on various benchmarks. 
Specifically, HGFormer consistently outperforms the recent SoTA counterparts~\cite{hassani2023neighborhood,liu2022convnet,liu2021swin} across four fundamental benchmarks in Fig. \ref{fig:performance}.
Totally, we conduct benchmarks including image classification (Sec. \ref{sec:cls}), object detection and instance segmentation (Sec. \ref{sec:det}), semantic segmentation (Sec. \ref{sec:seg}), pose estimation (Sec. \ref{sec:pos}) and weakly supervised semantic segmentation (Sec. \ref{sec:wsss}). 
Additional benchmarks about saliency detection and multi-label classification are presented in the supplementary material.
We provide the architecture and complexity analysis (Sec. \ref{sec:arc}) as well as the throughput estimation (Sec. \ref{sec:throughput}) of the proposed HGFormer. 
And we conduct extensive ablation (Sec. \ref{sec:ablation}) and visualization (Sec. \ref{sec:vis}) studies for effectiveness and interpretation of our contributions.

\setlength{\tabcolsep}{6pt}
\begin{table*}[t]
    \caption{
    \textbf{ImageNet-1K image classification performance.} 
    All models are trained with the input of $224^{2}$ and the standard practice.  \textbf{`C'} denotes ConvNets, \textbf{`T'} denotes Transformers, \textbf{`G'} denotes ViHGNN and ViGs and \textbf{`H'} denotes Hybrid architectures. 
    }
    \vspace{-5pt}
    \centering
    \begin{minipage}{.41\linewidth}
        \centering
        \resizebox{1\textwidth}{!}{
        \renewcommand\arraystretch{1.3}
        \fontsize{20.5}{20.8}\selectfont
        \begin{tabular}{lcccc}
            \toprule
            \textbf{Model} & \textbf{Type} & \textbf{Params (M)}  & \textbf{FLOPs (G)} & \textbf{Top1 (\%)}\\
            \midrule
            \textbf{T2T-ViT-7}~\cite{t2t2021yuan} \textit{\large ICCV'2021}       &  \textbf{T}  &  4  &   1.1 & 71.1 \\
            \textbf{TNT-T}~\cite{han2021transformer} \textit{\large NIPS'2021}    &  \textbf{T}  &  6  &   1.4 & 73.9 \\
            \textbf{DependencyViT-T}~\cite{dependency} \textit{\large CVPR'2023}  &  \textbf{T}  &  6  &   1.3 & 75.4 \\
            \textbf{ViG-T}~\cite{han2022vision} \textit{\large NIPS'2022}         &  \textbf{G}  &  7  &   1.3 & 73.9 \\
            \textbf{ViHGNN-T}~\cite{han2023vision} \textit{\large ICCV'2023}      &  \textbf{G}  &  8  &   1.8 & 74.3 \\
            \textbf{MViG-T}~\cite{munir2023mobilevig} \textit{\large CVPR'2023}   &  \textbf{G}  &  5  &   0.7 & 75.7 \\
            \rowcolor{lightgray}
            \textbf{HGFormer-T}    &  \textbf{H} & 5 & 1.2 & \textbf{75.8} \\
            \midrule
            \textbf{CoAtNet-0}~\cite{coatnet} \large NIPS'2021           &  \textbf{C}  &  25  &  4.2 & 81.6 \\
            \textbf{ConvNeXt-T}~\cite{liu2022convnet} \textit{\large CVPR'2022}   &  \textbf{C}  &  28  &  4.5 & 82.1 \\
            \textbf{FocalNet-T}~\cite{focalnet} \textit{\large NIPS'2022}         &  \textbf{C}  &  28  &  4.5 & 82.1 \\
            \textbf{HorNet-T}~\cite{hornet} \textit{\large NIPS'2022}             &  \textbf{C}  &  22  &  4.0 & 82.8 \\
            \textbf{SLaK-T}~\cite{slak} \textit{\large ICLR'2023}                 &  \textbf{C}  &  30  &  5.0 & 82.5 \\
            \textbf{Swin-T}~\cite{liu2021swin} \textit{\large ICCV'2021}          &  \textbf{T}  &  29  &  4.5 & 81.3 \\
            \textbf{NAT-T}~\cite{hassani2023neighborhood} \textit{\large CVPR'2023}  &  \textbf{T}  &  28  & 4.3 & 83.2 \\
            \textbf{EAPT-S}~\cite{lin2021eapt} \textit{\large TMM'2023}              &  \textbf{T}  &  39  &  6.5 & 83.0 \\
            \textbf{Dilate-S}~\cite{jiao2023dilateformer} \textit{\large TMM'2023}   &  \textbf{T}  &  21  &  4.8 & 83.3 \\
            \textbf{ViG-S}~\cite{han2022vision}  \textit{\large NIPS'2022}           &  \textbf{G}  &  27  &  4.6 & 82.1 \\
            \textbf{ViHGNN-S}~\cite{han2023vision} \textit{\large ICCV'2023}         &  \textbf{G}  &  23  &  5.6 & 81.5 \\
            \textbf{MViG-B}~\cite{munir2023mobilevig} \textit{\large CVPR'2023}      &  \textbf{G}  &  27  &  2.8 & 82.6 \\
            \rowcolor{lightgray}
            \textbf{HGFormer-S}    &  \textbf{H}  &  28  &   4.5 & \textbf{83.4} \\
            \bottomrule
        \end{tabular}
        }
    \end{minipage}%
    \begin{minipage}{.449\linewidth}
        \centering
        \resizebox{1\textwidth}{!}{
        \renewcommand\arraystretch{1.3}
        \fontsize{20.5}{20.8}\selectfont
        \begin{tabular}{lcccc}
            \toprule
            \textbf{Model} & \textbf{Type} & \textbf{Params (M)}  & \textbf{FLOPs (G)} & \textbf{Top1 (\%)}\\
            \midrule
            \textbf{ResNet-152}~\cite{resnet} \textit{\large CVPR'2016}              &  \textbf{C}  &  60  &  11.6 & 82.0 \\
            \textbf{CoAtNet-1}~\cite{coatnet} \large NIPS'2021              &  \textbf{C}  &  42  &  8.4 & 83.3 \\
            \textbf{ConvNeXt-S}~\cite{liu2022convnet} \textit{\large CVPR'2022}      &  \textbf{C}  &  50  &  8.7 & 83.1 \\
            \textbf{FocalNet-S}~\cite{focalnet}  \textit{\large NIPS'2022}           &  \textbf{C}  &  50  &  8.6 & 83.4 \\
            \textbf{HorNet-S}~\cite{hornet} \textit{\large NIPS'2022}                &  \textbf{C}  &  50  &  8.8 & 84.0 \\
            \textbf{SLaK-S}~\cite{slak} \textit{\large ICLR'2023}                    &  \textbf{C}  &  55  &  9.8 & 83.8 \\
            \textbf{T2T-ViT-24}~\cite{t2t2021yuan} \textit{\large ICCV'2021}         &  \textbf{T}  &  64  &  13.8 & 82.3 \\
            \textbf{TNT-B}~\cite{han2021transformer} \textit{\large NIPS'2021}       &  \textbf{T}  &  66  &  14.1 & 82.8 \\
            \textbf{Swin-S}~\cite{liu2021swin} \textit{\large ICCV'2021}             &  \textbf{T}  &  50  &  8.7 & 83.0 \\
            \textbf{UniFormer-B}~\cite{li2022uniformer} \textit{\large ICLR'2022}    &  \textbf{T}  &  50  &  8.3 & 83.9 \\
            \textbf{CSWin-B}~\cite{cswin} \textit{\large CVPR'2022}                  &  \textbf{T}  &  78  &  15.0 & 84.2 \\
            \textbf{NAT-S}~\cite{hassani2023neighborhood} \textit{\large CVPR'2023}  &  \textbf{T}  &  51  &  7.8 & 83.7 \\
            \textbf{Dilate-B}~\cite{jiao2023dilateformer} \textit{\large TMM'2023}   &  \textbf{T}  &  47  &  10.0 & 84.4 \\
            \textbf{ViG-B}~\cite{han2022vision} \textit{\large NIPS'2022}            &  \textbf{G}  &  52  & 8.9 & 83.1 \\
            \textbf{ViHGNN-B}~\cite{han2023vision} \textit{\large ICCV'2023}         &  \textbf{G}  &  52  & 10.7 & 83.4 \\
            \textbf{PVG-M}~\cite{wu2023pvg} \textit{\large MM'2023}                  &  \textbf{G}  &  52  & 10.7 & 83.4 \\
            \rowcolor{lightgray}
            \textbf{HGFormer-B}                             &  \textbf{H}  &  54  &   8.7 & \textbf{84.4} \\
            \bottomrule
        \end{tabular}
        }
    \end{minipage}%
    \vspace{-8pt}
    \label{apptab:imagenet_comp}
\end{table*}

\subsection{Benchmark Evaluation}
\subsubsection{\textbf{Image Classification}}
\label{sec:cls}
Image classification is the fundamental task in computer vision, designed to train a backbone network to assign input images to predefined labels, serving as the basis for various downstream tasks.

\textbf{Implementation \& Settings.} 
ImageNet-1K~\cite{deng2009imagenet} dataset is a large-scale 1000-class dataset that contains 1.28M training images and 50K validation images.
For classification experiments from scratch on ImageNet-1K, we implement our models through \verb|timm|~\cite{rw2019timm} package.
We compare four typical architectures: 
(\romannumeral1) \textbf{ConvNet}.
(\romannumeral2) \textbf{Transformer}.
(\romannumeral3) \textbf{ViHGNN \& ViG}.
(\romannumeral4) \textbf{Hybrid architecture}.
For fair comparison, the training details of the proposed HGFormer adhere to the standard practices \cite{liu2021swin,liu2022convnet}, which are extensively adopted in transformers and convnets.
Specifically, we train three HGFormer variants for 300 epochs by AdamW optimizer using a basic learning rate of $1.5\times10^{-3}, 1.25\times10^{-3}, 1\times10^{-3}$, a cosine scheduler and a weight decay with 20 epochs of linear warmup.
For data augmentation and regularization techniques, we employ most of the strategies applied in \cite{liu2021swin,liu2022convnet}, including RandomAugment, Mixup, RandomErasing, etc.
For larger models, the extent of stochastic depth increase progressively, \textit{i.e.}, 0.05, 0.1, 0.15 for HGFormer-T, -S, -B, respectively.
\textbf{Results.} We compare the performances among three widely adopted model sizes (around 5M, 25M and 50M parameters) in Tab.~\ref{apptab:imagenet_comp}. It can be seen that HGFormers consistently outperform the recent SoTA counterparts with similar parameters (Params.) and computation (FLOPs), especially improving Swin-T/S by +1.9\%/+1.1\%, ConvNeXt-T/S by +1.0\%/+1.0\% and NAT-M/S by +1.4\%/+0.4\%. 
HGFormers also outperforms the recent ViG and ViHGNN counterparts, especially improving ViG-T/S/B by +1.9\%/+1.1\%/+1.0\% and ViHGNN-S/-M by +0.2\%/+1.0\%. 
Across all the three model sizes, HGFormers exhibit excellent scalability, manifesting the advantages of the proposed hybrid architecture that combines the topology prior of hypergraph with the capabilities of transformer.

\setlength{\tabcolsep}{3pt}
\begin{table*}[t]
    \caption{\textbf{COCO object detection and instance segmentation performance.} 
    All models are trained with standard practice of \textbf{`3x'}.
    }
    \vspace{-2pt}
    \centering
    \begin{minipage}{.49\linewidth}
        \centering
        \resizebox{1\textwidth}{!}{
        \renewcommand\arraystretch{1.3}
        \fontsize{20.5}{20.8}\selectfont
        \begin{tabular}{lcc|ccc|ccc}
            \toprule
            \multicolumn{9}{c}{\textbf{\textit{Mask R-CNN - 3x schedule}}} \\
            \midrule
            \textbf{Backbone} & \textbf{Params (M)} & \textbf{FLOPs (G)} & \textbf{AP\textsuperscript{b}} & \textbf{AP\textsuperscript{b}\textsubscript{50}} & \textbf{AP\textsuperscript{b}\textsubscript{75}} & \textbf{AP\textsuperscript{m}} & \textbf{AP\textsuperscript{m}\textsubscript{50}} & \textbf{AP\textsuperscript{m}\textsubscript{75}} \\ 
            \midrule
            \textbf{Swin-T}~\cite{liu2021swin} \textit{\large ICCV'2021}     &  48 &  267 & 46.0 & 68.1 & 50.3 & 41.6 & 65.1 & 44.9 \\
            \textbf{ConvNeXt-T}~\cite{liu2022convnet} \textit{\large CVPR'2022}  &  48 &  262 & 46.2 & 67.0 & 50.8 & 41.7 & 65.0 & 44.9 \\
            \textbf{FocalNet-T}~\cite{focalnet} \textit{\large NIPS'2022}        &  49 &  268 & 48.0 & \textbf{69.7} & 53.0 & 42.9 & 66.5 & 46.1 \\
            \textbf{NAT-T}~\cite{hassani2023neighborhood} \textit{\large CVPR'2023}    &  48 &  258 & 47.7 & 69.0 & 52.6 & 42.6 & 66.1 & 45.9 \\
            \rowcolor{lightgray}
            \textbf{HGFormer-S}   & 48 &  262 & \textbf{48.2}   & \textbf{69.7} & \textbf{53.1} & \textbf{43.1} & \textbf{66.8} & \textbf{46.4} \\
            \midrule
            \textbf{Swin-S}~\cite{liu2021swin} \textit{\large ICCV'2021}     & 69 &  359 & 48.5 & 70.2 & 53.5 & 43.3 & 67.3 & 46.6 \\
            \textbf{FocalNet-S}~\cite{focalnet} \textit{\large NIPS'2022}    & 72 &  365 & 49.3 & \textbf{70.7} & 54.2 & 43.8 & \textbf{67.9} & 47.4 \\
            \textbf{NAT-S}~\cite{hassani2023neighborhood} \textit{\large CVPR'2023}   & 70 &  330 & 48.4 & 69.8 & 53.2 & 43.2 & 66.9 & 46.5 \\
            \rowcolor{lightgray}
            \textbf{HGFormer-B}          & 70 &  361 & \textbf{49.4} & \textbf{70.7} & \textbf{54.5} & \textbf{43.9} & \underline{67.7} & \textbf{47.5} \\
            \bottomrule
        \end{tabular}
        }
    \end{minipage}%
    \begin{minipage}{.49\linewidth}
        \centering
        \resizebox{1\textwidth}{!}{
        \renewcommand\arraystretch{1.3}
        \fontsize{20.5}{20.8}\selectfont
        \begin{tabular}{lcc|ccc|ccc}
            \toprule
            \multicolumn{9}{c}{\textbf{\textit{Cascade Mask R-CNN - 3x schedule}}} \\
            \midrule
            \textbf{Backbone} & \textbf{Params (M)} & \textbf{FLOPs (G)} & \textbf{AP\textsuperscript{b}} & \textbf{AP\textsuperscript{b}\textsubscript{50}} & \textbf{AP\textsuperscript{b}\textsubscript{75}} & \textbf{AP\textsuperscript{m}} & \textbf{AP\textsuperscript{m}\textsubscript{50}} & \textbf{AP\textsuperscript{m}\textsubscript{75}} \\ 
            \midrule
                \textbf{Swin-T}~\cite{liu2021swin} \textit{\large ICCV'2021}             &  86 &  745 & 50.4 & 69.2 & 54.7 & 43.7 & 66.6 & 47.3 \\
            \textbf{ConvNeXt-T}~\cite{liu2022convnet} \textit{\large CVPR'2022}          &  86 &  741 & 50.4 & 69.1 & 54.8 & 43.7 & 66.5 & 47.3 \\
            \textbf{FocalNet-T}~\cite{focalnet} \large NIPS'2022                &  87 &  751 & 51.5 & 70.1 & 55.8 & 44.6 & 67.7 & 48.4 \\
            \textbf{NAT-T}~\cite{hassani2023neighborhood} \textit{\large CVPR'2023}      &  85 &  737 & 51.4 & 70.0 & 55.9 & 44.5 & 67.6 & 47.9 \\
            \rowcolor{lightgray}
            \textbf{HGFormer-S}   & 86 &  733 & \textbf{51.8} & \textbf{70.6} & \textbf{56.3} & \textbf{44.8} & \textbf{68.0} & \textbf{48.4} \\
            \midrule
            \textbf{Swin-S}~\cite{liu2021swin} \textit{\large ICCV'2021}                 & 107 &  838 & 51.8 & 70.4 & 56.3 & 44.7 & 67.9 & 48.5 \\
            \textbf{ConvNeXt-S}~\cite{liu2022convnet} \textit{\large CVPR'2022}          & 108 &  827 & 51.9 & 70.8 & 56.5 & 45.0 & 68.4 & 49.1 \\
            \textbf{NAT-S}~\cite{hassani2023neighborhood} \textit{\large CVPR'2023}      & 108 &  809 & 52.0 & 70.4 & 56.3 & 44.9 & 68.1 & 48.6 \\
            \rowcolor{lightgray}
            \textbf{HGFormer-B}  & 106 &  830 & \textbf{52.5} & \textbf{71.2} & \textbf{57.1} & \textbf{45.4} & \textbf{68.8} & \textbf{49.3} \\
            \bottomrule
        \end{tabular}
        }
    \end{minipage}%
    \vspace{-2pt}
    \label{tab:object_detection}
\end{table*}

\setlength{\tabcolsep}{20pt}
\begin{table*}[t]
    \caption{\textbf{ADE20K semantic segmentation performance.} 
    All models are trained with the input of $512^{2}$ and the standard practice of \textbf{160k}.
    }
    \vspace{-5pt}
    \centering
    \begin{minipage}{.48\linewidth}
        \centering
        \resizebox{1\textwidth}{!}{
        \renewcommand\arraystretch{1.3}
        \fontsize{20.5}{20.8}\selectfont
        \begin{tabular}{lcc|cc}
            \toprule
            \multicolumn{5}{c}{\textbf{\textit{UperNet - 160K}}} \\
            \midrule
            \textbf{Backbone} & \textbf{Params (M)} & \textbf{FLOPs (G)} & \textbf{mIoU} & \textbf{+MS} \\ 
            \midrule
            \textbf{Swin-T}~\cite{liu2021swin} \textit{\large ICCV'2021}                     & 60 & 946 & 44.5 & 45.8 \\
            \textbf{ConvNeXt-T}~\cite{liu2022convnet} \textit{\large CVPR'2022}     & 60 & 939 & 46.0 & 46.7 \\
            \textbf{NAT-T}~\cite{hassani2023neighborhood} \textit{\large CVPR'2023}     & 58 & 934 & 47.1 & 48.4 \\
            \textbf{ViT-Adapter-S}~\cite{adapter} \textit{\large ICLR'2023}                  & 58 & - & 46.2  & 47.1  \\
            \rowcolor{lightgray}
            \textbf{HGFormer-S}                            & 58 & 931 & \textbf{47.4} & \textbf{48.9}  \\
            \midrule
            \textbf{Swin-S}~\cite{liu2021swin}  \textit{\large ICCV'2021}                    & 81 &  1040 & 47.6 & 49.5  \\
            \textbf{ConvNeXt-S}~\cite{liu2022convnet} \textit{\large CVPR'2022}     & 82 &  1027 & 48.7 & 49.6  \\
            \textbf{NAT-S}~\cite{hassani2023neighborhood} \textit{\large CVPR'2023}     & 82 &  1010 & 48.0 & 49.5 \\
            \textbf{ViT-Adapter-B}~\cite{adapter} \textit{\large ICLR'2023}                  & 134 &  - & 48.8 & 49.7  \\
            \rowcolor{lightgray}
            \textbf{HGFormer-B}                                                     & 83 &  1022 & \textbf{49.0} & \textbf{50.1}  \\
            \bottomrule
        \end{tabular}
        }
    \end{minipage}%
    \begin{minipage}{.514\linewidth}
        \centering
        \resizebox{1\textwidth}{!}{
        \renewcommand\arraystretch{1.3}
        \fontsize{20.5}{20.8}\selectfont
        \begin{tabular}{lcc|cc}
            \toprule
            \multicolumn{5}{c}{\textbf{\textit{Mask2Former - 160K}}} \\
            \midrule
            \textbf{Backbone} & \textbf{Params (M)} & \textbf{FLOPs (G)} & \textbf{mIoU} & \textbf{+MS} \\ 
            \midrule
            \textbf{ResNet-50}~\cite{resnet} \textit{\large CVPR'2016}                       & 44 & 71 & 47.2 & 49.2  \\
            \textbf{Swin-T}~\cite{liu2021swin} \textit{\large ICCV'2021}                     & 47 & 74 & 47.7 & 49.6 \\
            \textbf{NAT-T}~\cite{hassani2023neighborhood} \textit{\large CVPR'2023}     & 47 & 72 & 48.2 & 50.8 \\
            \rowcolor{lightgray}
            \textbf{HGFormer-S}                      & 46 & 69 & \textbf{48.9} & \textbf{51.3}  \\
            \midrule
            \textbf{ResNet-101}~\cite{resnet} \textit{\large CVPR'2016}                      & 63 &  90 & 47.8 & 50.1  \\
            \textbf{Swin-S}~\cite{liu2021swin} \textit{\large ICCV'2021}                     & 69 &  98 & 51.3 & 52.4  \\
            \textbf{NAT-S}~\cite{hassani2023neighborhood} \textit{\large CVPR'2023}     & 68 & 92 & 52.2 & 53.2 \\
            \textbf{TransNeXt-T}~\cite{transnext} \textit{\large CVPR'2024}                  & 47 & 74  & \textbf{53.4} & -   \\
            \rowcolor{lightgray}
            \textbf{HGFormer-B}               & 61 &  91 & \underline{52.8} & \textbf{53.5}  \\
            \bottomrule
        \end{tabular}
        }
    \end{minipage}%
    \vspace{-10pt}
\label{tab:semseg}
\end{table*} 

\subsubsection{\textbf{Object Detection and Instance Segmentation}}
\label{sec:det}
Object detection aims to identify predefined targets within images, determining their categories and locations.
Instance segmentation stands as a specialized form of object detection by outputting object masks instead of bounding boxes.
They are both the fundamental and challenging tasks for visual understanding.

\textbf{Implementation \& Settings.}
COCO 2017 \cite{lin2014microsoft} dataset contains 118K images for training, 5K images for validation and 20K images for testing. 
We benchmark the proposed HGFormer for object detection and instance segmentation on COCO 2017.
Specifically, we employ the ImageNet-1K pretrained HGFormer as the backbone of Mask R-CNN \cite{he2017mask} and Cascade Mask R-CNN \cite{zhao2017cascade} frameworks. 
Following standard practice in \cite{liu2021swin,liu2022convnet}, we implement our models through \verb|mmdetection|~\cite{chen2019mmdetection} package. 
\textbf{Settings.}
For fair comparison, the training details for object detection and instance segmentation follow the standard practices \cite{liu2021swin,liu2022convnet,hassani2023neighborhood} of $3\times$ schedule with 36 epochs.
Specifically, we employ AdamW optimizer with a basic learning rate of $1\times10^{-4}$ and a batch size of 16. 
The multi-scale training strategy is adopted and the shorter side of the image is resized within $[480, 800]$. In the testing phase, the shorter side of the input image is fixed to $800$ pixels.
\textbf{Results.} 
Object detection and instance segmentation results are presented in Tab.~\ref{tab:object_detection}.
With the two detection frameworks, HGFormers achieve considerable performance gains over the recent SoTA counterparts. Specifically, in Mask R-CNN, HGFormer-S gains +2.2AP\textsuperscript{b}/+1.5AP\textsuperscript{m} over Swin-T, +2.0AP\textsuperscript{b}/+1.4AP\textsuperscript{m} over ConvNeXt-T and +0.5AP\textsuperscript{b}/+0.5AP\textsuperscript{m} over NAT-T. In Cascade Mask R-CNN, HGFormer-S gains +1.4AP\textsuperscript{b}/+1.1AP\textsuperscript{m} over Swin-T, +1.4AP\textsuperscript{b}/+1.1AP\textsuperscript{m} over ConvNeXt-T and +0.4AP\textsuperscript{b}/+0.3AP\textsuperscript{m} over NAT-T.

\subsubsection{\textbf{Semantic Segmentation}}
\label{sec:seg}
Semantic segmentation aims to assign semantic categories to each pixel within images, which is the fundamental dense prediction tasks for precise visual content understanding.

\textbf{Implementation \& Settings.} 
ADE20K \cite{zhou2017scene} dataset contains 150 semantic categories, with 20K images for training and 2K images for validation.
We benchmark the proposed HGFormer for semantic segmentation on ADE20K. Specifically, we employ the ImageNet-1K pretrained HGFormer as the backbone of both UperNet \cite{xiao2018unified} and Mask2Former \cite{cheng2022masked} frameworks. 
Following standard practice in \cite{liu2021swin,liu2022convnet}, we implement our models through \verb|mmsegmentation|~\cite{mmseg2020} package.
For fair comparison, the training details for semantic segmentation on ADE20K \cite{zhou2017scene} follow the standard practices \cite{liu2021swin,liu2022convnet,hassani2023neighborhood} of 160K.
Specifically, we employ AdamW optimizer using a basic learning rate of $6\times10^{-5}$, a weight decay of 0.01, a batch size of 16, a linear scheduler with a linear warmup of 1,500 iterations. 
The size of the training images is resized and cropped to $512^2$, and the shorter side of the testing images is scale to $512$. \textbf{Results.} Semantic segmentation results are presented in Tab.~\ref{tab:semseg}.
With the two segmentation frameworks, HGFormers significantly outperform the recent SoTA counterparts. 
Specifically, in UperNet, HGFormers surpass Swin-T/S by +3.1/+0.6, ConvNeXt-T/S by +2.2/+0.5 and NAT-T/S by +0.5/+0.6 multi-scale mIoU.
In Mask2Former, HGFormers surpass Swin-T/S by +1.7/+1.1 and NAT-T/S by +0.5/+0.3 multi-scale mIoU.

\subsubsection{\textbf{Pose Estimation}}
\label{sec:pos}
Pose estimation aims to localize predefined keypoints, known as body joints, within images. This is a challenging task for image region understanding and relationship modeling.

\setlength{\tabcolsep}{3pt}
\begin{table}[t]
    \centering
    \caption{
    \textbf{COCO pose estimation performance.} 
    All models are trained with the input of $256\times 192$.
    }
    \vspace{-5pt}
    \resizebox{0.48\textwidth}{!}{
        \begin{tabular}{lcc|ccccc}
            \toprule
            \textbf{Backbone} & \textbf{Params (M)} & \textbf{FLOPs (G)} & \textbf{AP}  & \textbf{AP$^{50}$} & \textbf{AP$^{75}$} & \textbf{AR}   \\
            \midrule
            \textbf{HRNet-W32}~\cite{hrnet} \textit{\tiny CVPR'2019}        & 28.5   & 7.1   & 74.4      & 90.5      & 81.9      & 78.9      \\
            \textbf{Swin-T}~\cite{liu2021swin} \textit{\tiny ICCV'2021}     & 32.8   & 6.1   & 72.4      & 90.1      & 80.6      & 78.2      \\
            \textbf{ConvNeXt-T}~\cite{liu2022convnet} \textit{\tiny CVPR'2022}    & 33.1  & 5.5   & 73.2      & 90.0      & 80.9      & 78.8      \\
            \rowcolor{lightgray}
            \textbf{HGFormer-S}    & 32.6   & 6.0   & \textbf{74.8}  & \textbf{91.9}  & \textbf{82.3} & \textbf{80.1}      \\
            \bottomrule
        \end{tabular}
    }
    \vspace{-8pt}
    \label{tab:coco_pose}
\end{table}
\begin{figure}[t]
    \centering
    \includegraphics[width=0.96\columnwidth]{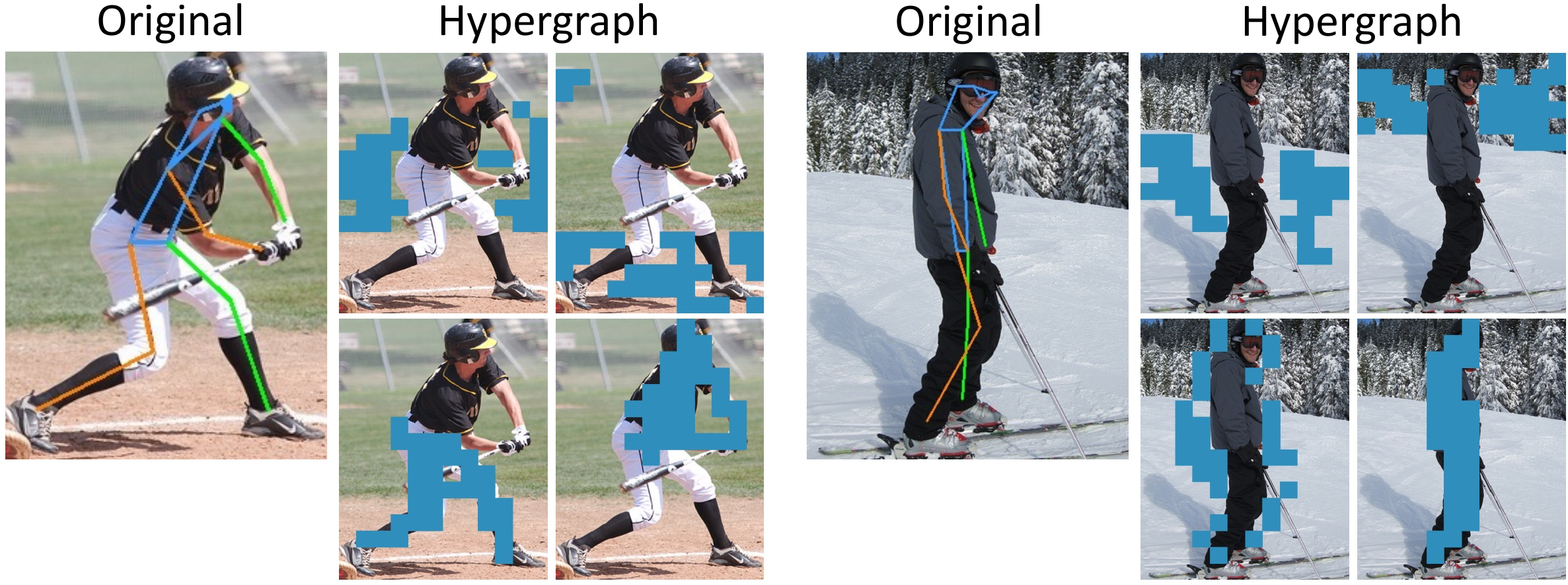}
    \vspace{-6pt}
    \caption{
    \textbf{Pose and Hypergraph Visualization on COCO Pose Estimation (Heatmap-Based).} 
    The masked regions denote the semantically relevant nodes connected within the same hyperedge.
    Both in the human body and in the background, regions with similar semantics are effectively grouped.
    }
    \vspace{-8pt}
    \label{fig:vis_poses}
\end{figure}

\textbf{Implementation \& Settings.} 
COCO 2017 \cite{lin2014microsoft} pose estimation dataset contains over 250K instance annotations for 17 predefined human keypoints.
We benchmark our HGFormers for pose estimation on COCO 2017. Specifically, we employ the ImageNet-1K pretrained HGFormer as the backbone of SimpleBaseline \cite{he2016deep} framework. Following the standard practices \cite{hrnet}, we implement our models through 
\verb|mmpose|~\cite{mmpose2020} 
package and adopt the same training and evaluation settings. Specifically, we employ AdamW optimizer with a base learning rate of $8\times10^{-4}$, a weight decay of 0.05 and a batch size of 256, respectively. Totally, we train the model for 210 epochs. \textbf{Results.} Pose estimation results are presented in Tab. \ref{tab:coco_pose}. It can be seen that HGFormer achieves performance improvements over the recent counterparts. Specifically, HGFormer-S yield +1.6/+1.5/+1.2 AP and +1.2 AR improvements over ConvNeXt-T. 
The results obtained solely based on the simple framework demonstrates the impact of the hypergraph concept on regional context and spatial topology, contributing to the pose estimation.
Additionally, we visualize the pose estimation results and the corresponding hypergraph topologies in Fig.~\ref{fig:vis_poses}.

\subsubsection{\textbf{Weakly Supervised Semantic Segmentation}}
\label{sec:wsss}
Weakly Supervised Semantic Segmentation (WSSS) aims to perform semantic segmentation with weak labels like image-level labels instead of pixel-level annotations.

\setlength{\tabcolsep}{9pt}
\begin{table}[t]
    \centering
    \caption{
    \textbf{PASCAL VOC 2012 weakly supervised semantic segmentation performance.} 
    All models are single-staged methods and supervised with image-level labels.
    }
    \vspace{-5pt}
    \resizebox{0.49\textwidth}{!}{
        \begin{tabular}{lc|cc}
            \toprule
            \textbf{Method} & \textbf{Backbone} & \textbf{mIoU (\textit{val})} & \textbf{mIoU (\textit{test})}   \\
            \midrule
            \textbf{1Stage}~\cite{1Stage} \textit{\tiny CVPR'2020} & \textbf{ResNet-38}   & 62.7   & 64.3  \\
            \textbf{AFA}~\cite{afa} \textit{\tiny CVPR'2022} & \textbf{MiT-B1}  & 66.0   & 66.3            \\
            \textbf{ToCo}~\cite{toco} \textit{\tiny CVPR'2023} & \textbf{ViT-B}  & 71.1   & 72.2           \\
            \rowcolor{lightgray}
            \textbf{Ours}  & \textbf{HGFormer-B} & \textbf{73.8}   & \textbf{74.4}      \\
            \bottomrule
        \end{tabular}
    }
    \vspace{-6pt}
    \label{tab:wsss}
\end{table}

\begin{figure}[t]
    \centering
    \includegraphics[width=1\columnwidth]{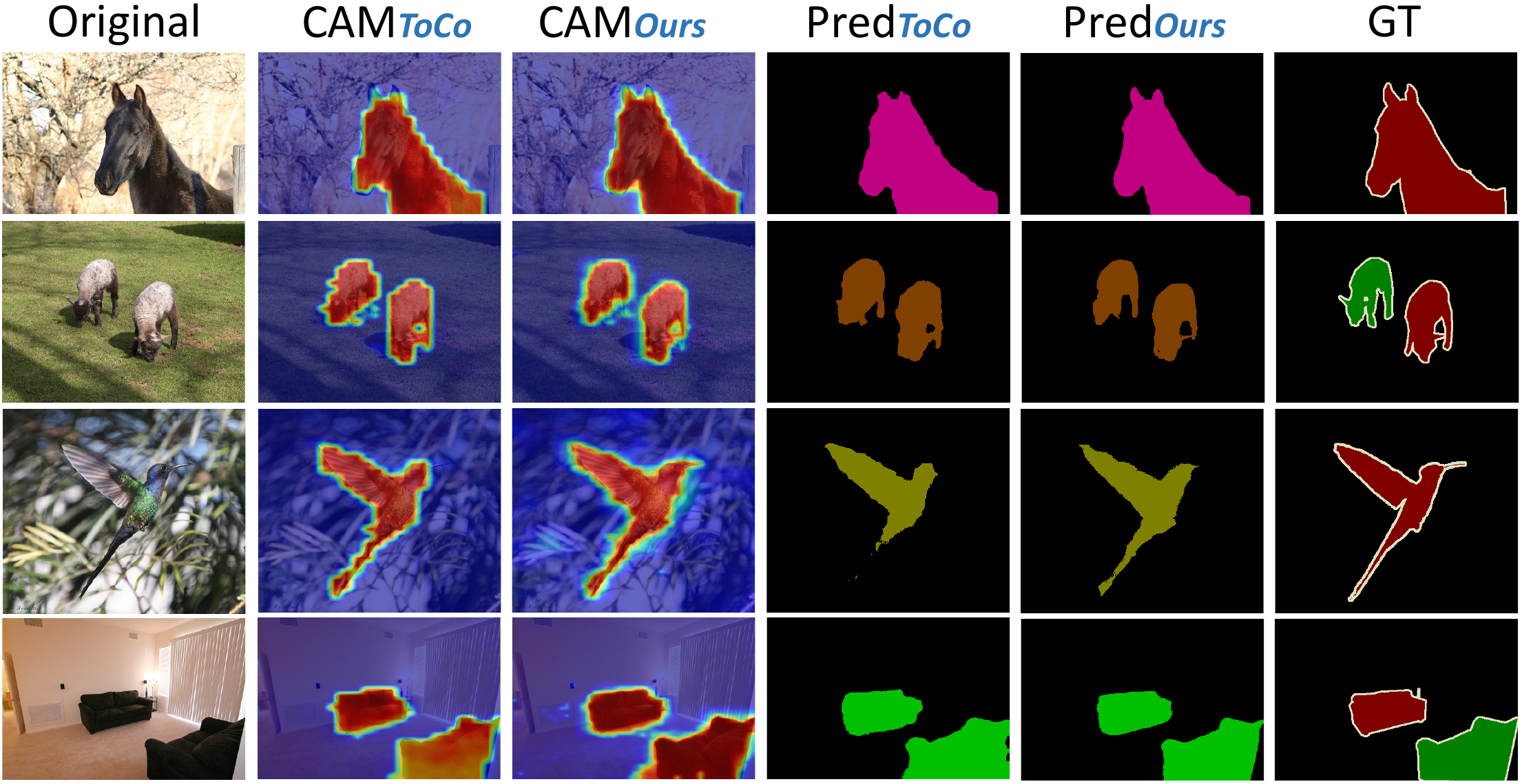}
    \vspace{-18pt}
    \caption{
    \textbf{Weakly Supervised Semantic Segmentation on PASCAL VOC 2012.} 
    We employ the proposed HGFormer as the backbone of ToCo~\cite{toco} framework to replace the original ViT~\cite{vit}. 
    \textit{
    Zoom in for better view.
    }
    }
    \vspace{-8pt}
    \label{fig:vis_wsss}
\end{figure}

\textbf{Implementation \& Settings.} 
PASCAL VOC 2012 \cite{voc} dataset contains 10582 images for training, 1449 images for validating and 1456 images for testing. 
We benchmark our HGFormers for weakly supervised semantic segmentation on PASCAL VOC 2012. Specifically, we employ the ImageNet-1K pretrained HGFormer as the backbone of ToCo~\cite{toco} framework to replace the original ViT~\cite{vit}. 
During training, we only use image-level labels. 
We employ AdamW optimizer, during which the learning rate linearly increases to $6\times10^{-5}$ in the first 1500 iterations and decays with a polynomial scheduler for later iterations. 
The warm-up and decay rates are set as $1\times10^{-6}$ and 0.9, respectively.
\textbf{Results.} Weakly supervised semantic segmentation results are presented in Tab. \ref{tab:wsss}. HGFormer achieves performance improvements over the recent counterparts. Specifically, Our method yield +2.7 mIoU on val set and +2.2 mIoU on test set over the baseline ToCo. 
WSSS works with image-level labels generally derive Class Activation Map (CAM) as pseudo labels, process them with refinement methods, and use to train segmentation models. 
Existing works usually concentrate on elaborating various frameworks, while overlooking the fact that the CAM-recognized regions are highly backbone-dependent.
Based on the observations in Fig.~\ref{fig:feat_intro} and Fig.~\ref{fig:vis_feat}, we believe that the capability of backbone is equally important.
The visualization in Fig.~\ref{fig:vis_wsss} demonstrates that the proposed HGFormer helps to generate more accurate CAMs and predictions, especially producing detailed and smooth boundaries.

\subsection{Architecture and Complexity Analysis.}
\label{sec:arc}

\begin{figure}[t]
    \centering
    \includegraphics[width=0.86\columnwidth]{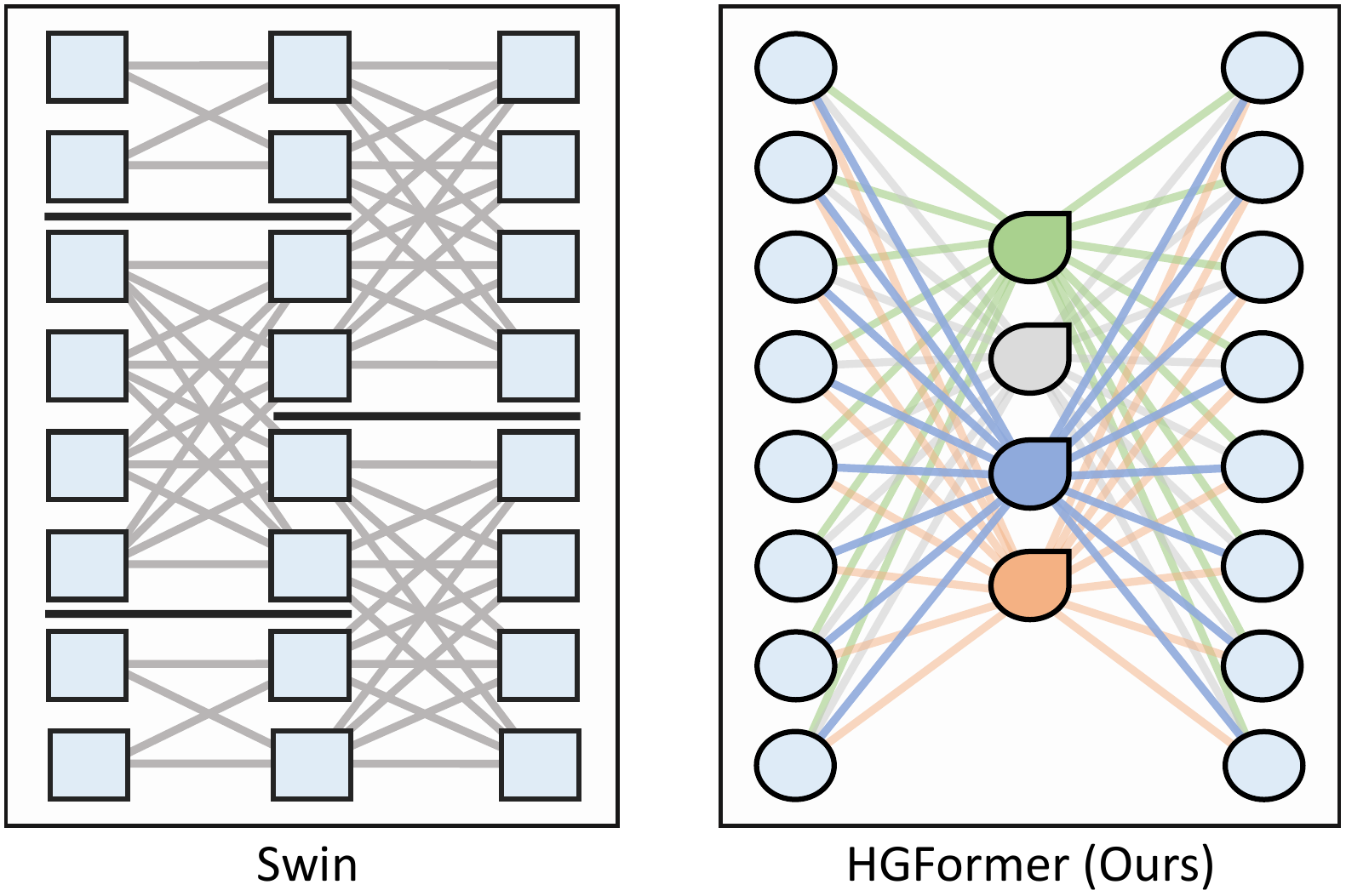}
    \vspace{-6pt}
    \caption{
    \textbf{Architecture and complexity comparisons between the proposed HGFormer and Swin~\cite{liu2021swin}.} Swin restricts attention to local windows, with shifts for cross-window interactions. The proposed HGFormer is configured with node-hyperedge-node messaging for relational reasoning, where the middle hyperedge tokens in different colors signify different local topologies. 
    }
    \vspace{-8pt}
    \label{fig:arc_comp}
\end{figure}

We compare architecture and complexity between the proposed HGFormer and the popular window-based method Swin in Fig.~\ref{fig:arc_comp}: (\romannumeral 1) Compared to windows with fixed locations and regular shapes, the proposed hypergraph topology preserves regional context and spatial topology with arbitrary locations and irregular shapes, better adapting to local variations within image; 
(\romannumeral 2) Compared to implicit modeling, the proposed HGFormer performs relational reasoning based on the node-hyperedge-node messaging, achieving higher-order modeling. 
Additionally, we analyze the complexity of the both models. 
The attention mechanism remains the main operation of both models, consuming most of the parameters and computation.
HGFormer maintains similar complexity to Swin while achieving better accuracy across various visual benchmarks.
The main operations of the two models share similar complexity $O(nck)$, where $n$ denotes the number of pixel / node tokens, $c$ denotes the embedding dimension, and $k$ denotes window size / hyperedge number.

\subsection{Throughput Estimation.}
\label{sec:throughput}

\setlength{\tabcolsep}{4pt}
\begin{table}[t]
    \centering
    \caption{
    \textbf{Inference throughput performance on ImageNet.}  
    }
    \vspace{-4pt}
    \resizebox{0.48\textwidth}{!}{
        \begin{tabular}{lc|ccc|c}
            \toprule
            \multirow{2}{*}{\textbf{Model}}  & \multirow{2}{*}{\textbf{Type}} & \textbf{Params}  & \textbf{FLOPs}  & \textbf{Throughput} & \textbf{Top1} \\
            & & \textbf{(M)} & \textbf{(G)} & \textbf{(image/s)} & \textbf{(\%)} \\
            \midrule
            \textbf{ConvNeXt-T}~\cite{liu2022convnet} \textit{\tiny CVPR'2022}    &  \textbf{C}  &  28  &  4.5 & \textbf{774.7} & 82.1 \\
            \textbf{Swin-T}~\cite{liu2021swin} \textit{\tiny ICCV'2021}            &  \textbf{T}  &  29  &  4.5 & 755.2 & 81.3 \\
            \textbf{NAT-S}~\cite{hassani2023neighborhood} \textit{\tiny CVPR'2023} &  \textbf{T}  &  28  &  4.3 & 693.2 & 83.2 \\
            \rowcolor{lightgray}
            \textbf{HGFormer-S}     &  \textbf{H}  &  28  &   4.5  & 770.2 & \textbf{83.4} \\
            \midrule
            \textbf{ConvNeXt-S}~\cite{liu2022convnet} \textit{\tiny CVPR'2022}     &  \textbf{C}  &  50  &  8.7 & \textbf{447.1} & 83.1 \\
            \textbf{Swin-S}~\cite{liu2021swin} \textit{\tiny ICCV'2021}            &  \textbf{T}  &  50  &  8.7 & 436.9 & 83.0 \\
            \textbf{NAT-S}~\cite{hassani2023neighborhood} \textit{\tiny CVPR'2023} &  \textbf{T}  &  51  &  7.8 & 432.9 & 83.7 \\
            \rowcolor{lightgray}
            \textbf{HGFormer-B}     &  \textbf{H}  &  54  &   8.7 & 442.3 & \textbf{84.4} \\
            \bottomrule
        \end{tabular}
    }
    \vspace{-12pt}
    \label{tab:throughput}
\end{table}

For throughput estimation, we benchmark HGFormers for image classification on ImageNet~\cite{deng2009imagenet}.
Following standard practice~\cite{deit,liu2021swin}, inference throughput in Tab.~\ref{tab:throughput} is measured through \verb|timm|~\cite{rw2019timm} package on a single NVIDIA V100 GPU with the input of $224^{2}$. Compared to the recent transformer Swin~\cite{liu2021swin} and NAT~\cite{hassani2023neighborhood} with similar Params and FLOPs, HGFormer achieves the highest throughput and accuracy. 
Although hypergraph construction consumes computation, the number of hyperedges is spatially reduced~\cite{wang2021pyramid, wang2022pyramid2} before clustering, significantly alleviating computational burden.
Thus, HGFormer achieves competitive throughput performance.

\subsection{Ablation Study} 
\label{sec:ablation}
To evaluate the impact of key components and strategies within the proposed HGFormer, we conduct several ablation studies based on HGFormer-T on ImageNet. 

\setlength{\tabcolsep}{15pt}
\begin{table}[t]
    \centering
    \caption{\textbf{Ablation comparison in the number of neighbors.} 
    }
    \vspace{-5pt}
    \resizebox{0.48\textwidth}{!}{
    \begin{tabular}{l|c|c}
        \toprule
        \textbf{Algorithm} & \textbf{Num. of neighbors $K$} & \textbf{Top-1 (\%)} \\
        \midrule
        \textbf{KNN} & 32, 16, 8, 4    & 74.0        \\
        \textbf{KNN} & 64, 32, 16, 8   & 74.0        \\
        \textbf{KNN} & 128, 64, 32, 8  & 74.4        \\
        \midrule
        \textbf{DPC-KNN} & 32, 16, 8, 4    & 74.3        \\
        \textbf{DPC-KNN} & 64, 32, 16, 8   & 74.9        \\
        \textbf{DPC-KNN} & 128, 64, 32, 8  & 75.2        \\
        \midrule
        \textbf{K-Means} & 32, 16, 8, 4    & 74.5        \\
        \textbf{K-Means} & 64, 32, 16, 8   & 74.8        \\
        \textbf{K-Means} & 128, 64, 32, 8  & 75.0        \\
        \midrule
        \rowcolor{lightgray}
        \textbf{CS-KNN (Ours)} & 32, 16, 8, 4    & 74.8        \\
        \rowcolor{lightgray}
        \textbf{CS-KNN (Ours)} & 64, 32, 16, 8   & 75.2        \\
        \rowcolor{lightgray}
        \textbf{CS-KNN (Ours)} & 128, 64, 32, 8  & \textbf{75.8}        \\
        \bottomrule
    \end{tabular}
    }
    \vspace{-12pt}
    \label{tab:algo}
\end{table}
\begin{figure}[t]
    \centering
    \includegraphics[width=0.98\columnwidth]{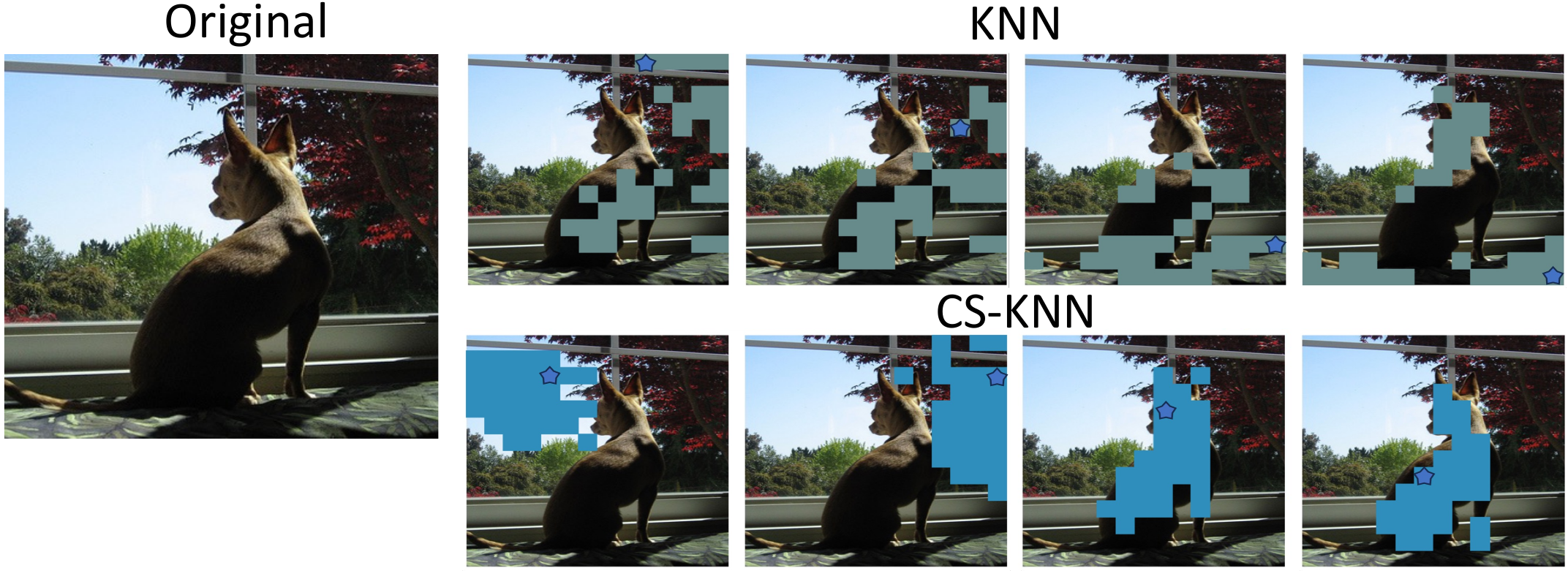}
    \vspace{-7pt}
    \caption{
    \textbf{Hypergraph comparison between the proposed CS-KNN and KNN.} The stars denote clustering centers and the nodes with dependencies are visualized by the same color.
    Using KNN, clustering centers like noise and isolated pixels introduce semantic confusion to their connected hyperedges.
    The proposed CS-KNN utilizes the class token to identify the informative tokens as sampling centers to construct robust hypergraph topologies.
    }
    \vspace{-12pt}
    \label{fig:knn}
\end{figure}
\setlength{\tabcolsep}{11pt}
\begin{table}[t]
    \centering
    \caption{
    \textbf{Distance Function.}  
    }
    \vspace{-5pt}
    \resizebox{0.49\textwidth}{!}{
    \renewcommand\arraystretch{1.}
    \begin{tabular}{l|c|c|c|c}
        \toprule
        \textbf{Function} & \textbf{Euclidean} & \textbf{Cosine} & \textbf{Softmax} & \textbf{Dot} \\
        \midrule
        \textbf{Top-1} (\%) & 74.7 & 74.8  & 75.2 & \textbf{75.8}    \\
        \bottomrule
    \end{tabular}
    }
    \vspace{-10pt}
    \label{tab:dist}
\end{table}
\setlength{\tabcolsep}{7pt}
\begin{table}[t]
    \centering
    \caption{
    \textbf{Ablation of Attention and hierarchical architecture.} 
    }
    \vspace{-5pt}
    \resizebox{0.48\textwidth}{!}{
    \begin{tabular}{l|cc|c}
        \toprule
        \textbf{Variant} & \textbf{Params}  & \textbf{FLOPs} & \textbf{Top-1 (\%)} \\
        \midrule
        \textbf{HGFormer-T with \textit{vanilla attention}}   &  4.9 M  & 1.0 G & 74.0 \\
        \textbf{HGFormer-T with \textit{single stage}}   &  5.1 M  & 1.2 G & 74.2 \\
        \textbf{HGFormer-T}         &  5.1 M  & 1.2 G & \textbf{75.8} \\
        \bottomrule
    \end{tabular}
    }
    \vspace{-8pt}
    \label{tab:hgattn}
\end{table}

\subsubsection{\textbf{Effect of CS-KNN algorithm}}
All the four methods KNN, DPC-KNN, K-Means, and CS-KNN can be adopted for hypergraph construction. 
Advanced beyond the traditional clustering algorithms, the proposed CS-KNN integrates the class token~\cite{liang2022not,fayyaz2022adaptive} in vision transformer for semantic guidance in clustering.
Tab.~\ref{tab:algo} shows that CS-KNN performs best with little overhead.
Aside from quantitative comparisons, we visually compared the differences between the proposed CS-KNN and KNN in Fig.~\ref{fig:knn}.
By contrast, CS-KNN utilizes the class token to identify the informative tokens as sampling centers, filtering out a substantial amount of noise and outliers to construct robust hypergraph topologies.
Additionally, we evaluate various functions for token distance in Tab.~\ref{tab:dist}, and normalized dot product emerges as the best performer.

\subsubsection{\textbf{Hyperparameter of Hypergraph}}
Following spatial reduction mechanism in~\cite{wang2021pyramid, wang2022pyramid2}, the reduction ratio of the hyperedge number $N_e$ at each stage is set as [$\frac{1}{8}$,$\frac{1}{4}$,$\frac{1}{2}$,$1$] by the proposed center sampling strategy to reduce computation and redundancy. 
The hyperedge degree K is a hyperparameter that controls the neighborhood and regional scope of hypergraph. 
Too few neighbors restrict receptive field, while too many neighbors lead to over-smoothing. 
Empirical evidence shows that as stages deepen, neural networks gradually expand receptive field and enhance abstract representation. 
To dynamically adapt to this incremental change, we set the principle that the proportion of neighbors in all nodes increases gradually.
Tab.~\ref{tab:algo} shows that setting the degree K at each stage to [128,64,32,8] performs the best.

\subsubsection{\textbf{Effect of topology-aware HGA}}
Advanced beyond the vanilla attention in ViT, the proposed topology-aware HGA integrates hypergraph topology as perceptual indications to guide the aggregation of global and unbiased information.
To verify whether the perceptual indications help improve performance, we set a comparison model with vanilla attention that contains no hypergraph information. 
Tab.~\ref{tab:hgattn} shows that the proposed HGA brings +1.8\% improvement.

\begin{figure*}[!t]
    \centering
    \includegraphics[width=0.7\textwidth]{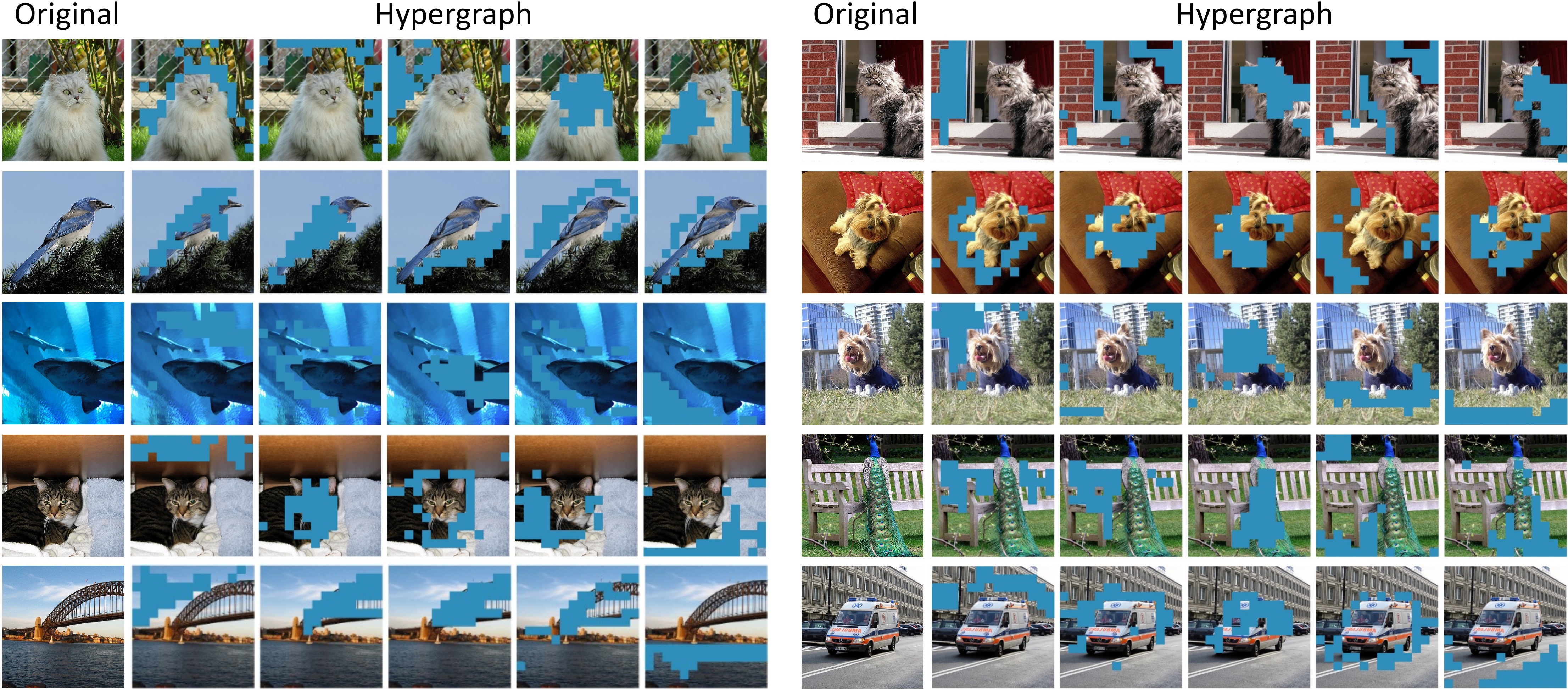}
    \vspace{-6pt}
    \caption{
    \textbf{Hypergraph Visualization on ImageNet.} 
    Moving beyond grid or sequence, input image or feature map is transformed into hypergraph by the proposed CS-KNN algorithm.
    The masked regions denote the semantically relevant nodes connected within the same hyperedge. 
    To avoid confusion caused by overlapping, we present hyperedges separately.
    The proposed hypergraph effectively preserves dependencies of the representative regions, outlining objects or scenes with arbitrary locations and irregular shapes.
    \textit{
    Zoom in for better view.
    }
    }
    \label{fig:vis_H}
    \vspace{-8pt}
\end{figure*}

\begin{figure*}[!t]
    \centering
    \includegraphics[width=0.7\textwidth]{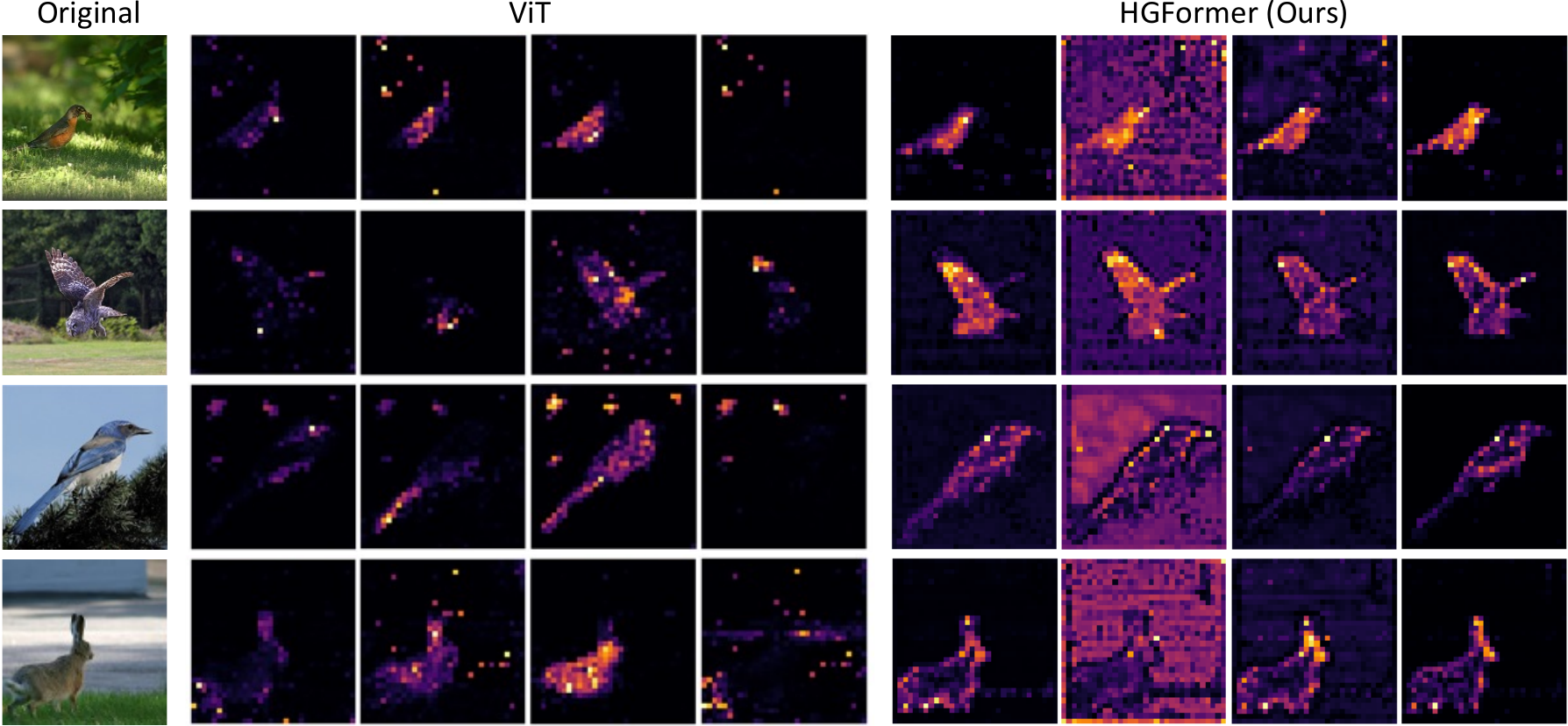}
    \vspace{-6pt}
    \caption{
    \textbf{Attention Visualization in different methods on ImageNet.} 
    The vanilla attention in ViT fails to accurately outline the target objects, leading to fragments and artifacts manifested as abnormally high values for certain tokens. Some specific heads even didn't realize the target objects.
    The proposed topology-aware HGA in HGFormer perfectly outlines target objects, significantly distinguishing between background and foreground.
    By contrast, the HGA exhibits superior capacity in regional context and spatial topology.
    \textit{
    Zoom in for better view.
    }
    }
    \label{fig:vis_attn}
    \vspace{-10pt}
\end{figure*}


\subsubsection{\textbf{Effect of Hypergraph Messaging}}
Advanced beyond the implicit modeling in ViT, the proposed HGFormer block is configured with hypergraph messaging for relational reasoning, involving node-hyperedge-node subroutines.
To verify whether the reasoning modeling and hierarchical architecture help improve performance, we design a comparison model with single-stage of node-hyperedge. 
Tab.~\ref{tab:hgattn} shows that the proposed architecture brings +1.6\% improvement.

\subsection{Visualization}
\label{sec:vis}


\begin{figure*}[!t]
    \centering
    \includegraphics[width=0.7\textwidth]{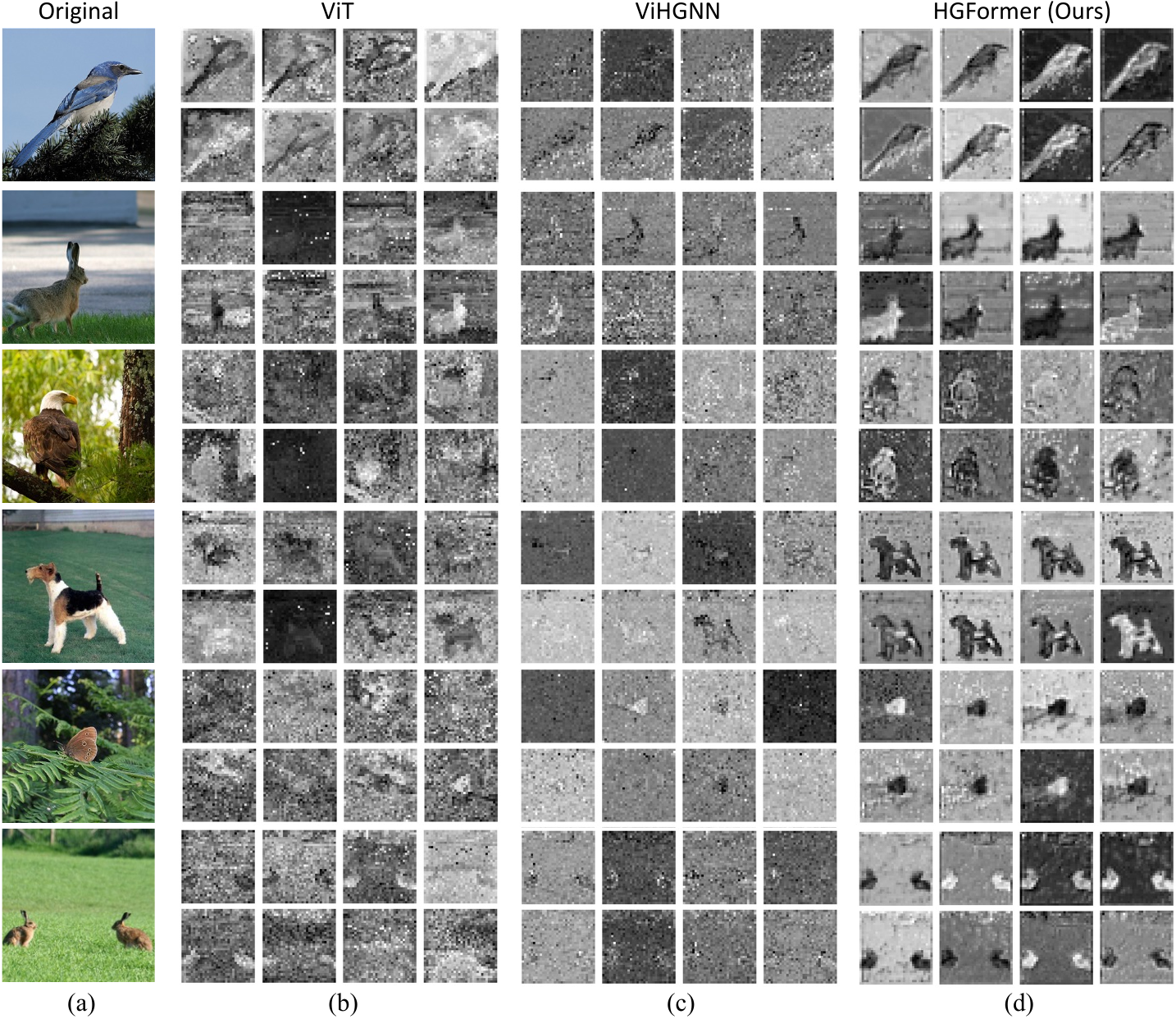}
    \vspace{-10pt}
    \caption{
    \textbf{Feature Visualization in different methods on ImageNet}. 
    (a) Original Image. 
    (b) ViT. 
    (c) ViHGNN.
    (d) HGFormer(Ours).
    ViT blends the foreground with the background ambiguously. 
    ViHGNN distinguishes between foreground and background, but its portrayals of objects is rough and unclear. 
    HGFormer(Ours) significantly highlights the foreground and suppresses the background, achieving a detailed depiction of objects.
    Note that the parameters and computation cost in these counterparts are similar. 
    \textit{
    Zoom in for better view.
    }
    }
    \label{fig:vis_feat}
    \vspace{-10pt}
\end{figure*}

To illustrate the inner workings of the proposed HGFormer, we introduce three visualization studies from the following aspects: hypergraph, attention and feature.

\subsubsection{\textbf{Hypergraph Visualization}}
We transform the input feature map into hypergraph topology by the proposed CS-KNN algorithm for regional context and spatial topology.
In Fig.~\ref{fig:vis_H}, the original image and five hyperedges are visualized, where the masked regions denote the semantically relevant nodes connected in the same hyperedge.
As can be seen, each hyperedge selects the most semantically relevant regions, outlining objects or scenes with arbitrary locations and irregular shapes. 
Thus, the topology helps to capture local variations, identify discriminative characters and aid in representation learning.

\subsubsection{\textbf{Attention Visualization}}
The vanilla attention in ViT possesses permutation invariance and fully-connected interaction, lacking an inductive bias in local structure and spatial correlation. 
Compared to the implicit modeling, the proposed topology-aware HGA integrates hypergraph topology as perceptual indications to guide aggregation of global and unbiased inforh ghgmation.
We visually compare the attention maps between vanilla attention and HGA in Fig.~\ref{fig:vis_attn}.
By contrast, the four heads in HGA perfectly outline target objects, distinguish between background and foreground, exhibiting superior capacity in regional context and spatial topology.

\subsubsection{\textbf{Feature Visualization}}
The permutation invariance and the fully-connected interaction in ViT disrupt regional context and spatial topology, leading to ambiguous scene understanding in Fig.~\ref{fig:vis_feat}b.
The strict structural inductive bias of cascaded HGConvs in ViHGNN causes locality, over-smoothing and error accumulation, manifesting as rough portrayals in Fig.~\ref{fig:vis_feat}c.
We propose to incorporate the topology perception of HGConv as perceptual indications and the global understanding of transformer for contextual refinement.
As visualized in Fig.~\ref{fig:vis_feat}d, we develop an effective and unitive representation, achieving distinct and detailed scene depiction.

\section{Conclusion \& Limitation}
\label{sec:conclusion}
In this work, we propose a topology-aware vision transformer called HyperGraph Transformer (HGFormer): (\romannumeral 1) The proposed Center Sampling K-Nearest Neighbors (CS-KNN) algorithm employs the class token for semantic guidance in clustering; (\romannumeral 2) The proposed topology-aware HGA introduces the topology perception of HGConv as perceptual indication and the global understanding of Transformer for contextual refinement; (\romannumeral 3) Empirical experiments show the competitive performances of HGFormers. 
Extensive ablation and visualization studies provide comprehensive interpretation of our contributions. 
The potential limitations of HGFormer: 
(\romannumeral 1) The hypergraph construction requires consideration of multiple hyperparameters, making it challenging.
(\romannumeral 2) For various downstream tasks, input resolution and content vary significantly, requiring fine-grained adjustments for hypergraph structure to accommodate input variations.
The limitations are ubiquitous in existing (hyper)graph-based methods and improving them would be a worthy research.

\footnotesize
\bibliographystyle{IEEEtran}
\bibliography{arxiv}

}
\end{document}